\ifdefined\pdfminorversion
\fi

\documentclass[10pt,twocolumn,letterpaper]{article}

\usepackage[pagenumbers]{cvpr}
\usepackage{silence}

\usepackage{caption}   
\usepackage{multirow}  
\usepackage{array}     

\definecolor{cvprblue}{rgb}{0.21,0.49,0.74}
\usepackage[pagebackref,breaklinks,colorlinks,allcolors=cvprblue]{hyperref}

\title{Read Together, Answer Apart: How Earlier Questions Help Vision-Language Models Look}

\author{
Xinyi Guo$^{1,*}$\qquad Kaile Huang$^{1,*}$\qquad Xitong Ling$^{2,*}$\qquad Mingyi He$^{3}$\\[2pt]
Chende Zheng$^{4}$\qquad Jianyu Xu$^{1}$\qquad Chao He$^{1,\dagger}$\qquad Dongqing Song$^{1,\dagger}$\\[5pt]
{\normalsize $^{1}$South China Normal University \qquad $^{2}$Tsinghua University}\\
{\normalsize $^{3}$Beijing University of Chemical Technology \qquad $^{4}$Xi'an Jiaotong University}\\[3pt]
{\small\fontfamily{cmtt}\selectfont chaohe@m.scnu.edu.cn \quad songdongqing@m.scnu.edu.cn}
}

\begin{document}
\maketitle
\begingroup
\renewcommand{\thefootnote}{}
\footnotetext[0]{$^{*}$Equal contribution.\quad $^{\dagger}$Corresponding authors.}
\endgroup
\raggedbottom

\begin{abstract}
Vision-language models (VLMs) are often asked several questions about the same image in a single prompt (\emph{co-asking}), yet existing evaluations report conflicting accuracy gains and hallucinations at the prompt level without tracking how individual answers change.
By pairing each question's single-question and co-asked answers across $22$ open-weight VLMs ($1\text{B}$--$35\text{B}$), we find that erratic net shifts reflect two opposing effects: on Densely Captioned Images (DCI), every model repairs $15.4\%$--$29.8\%$ of its single-question errors under co-asking, while the share of correct answers turned wrong falls from $30.8\%$ on weaker models to $4.7\%$ on stronger ones.
Across all $14$ models supporting custom attention masks, we show through attention readouts and layer-window interventions that earlier same-image questions read visual information into their question tokens in the early-to-middle layers during prefill, helping later questions concentrate image attention onto their queried targets.
This visual help accumulates across the scene even across non-overlapping regions, whereas blocking earlier questions from attending to the image lowers both target localization and accuracy in all $14$ models.
While cross-question help is established during shared question reading, part of the co-asking damage enters later during answer decoding when the decoder attends to other question stems.
We therefore introduce \emph{Read Together, Answer Apart}, a training-free decoding mask that keeps shared prefill intact and isolates each answer from other stems, improving co-asking accuracy on $12$ of $14$ models on DCI (up to $+4.71$ percentage points, closing $70\%$ of the pooled gap to asking alone) and on $10$, $8$, and $5$ of the $14$ models on WebUI, FinTabNet, and GQA (up to $+11.4$, $+5.3$, and $+1.1$ points).
\end{abstract}

\section{Introduction}
\label{sec:intro}

\begin{figure}[t]
\centering
\includegraphics[width=\columnwidth]{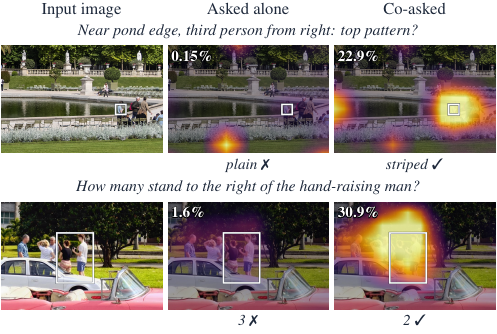}
\caption{\textbf{Earlier questions help a later question find its target.} Two repaired DCI examples on Gemma-4-12B (layers $17, 23, 29$ of $48$ at the final question token; answers abbreviated). Each pair shares a display scale; white boxes mark targets and percentages report unsmoothed in-box shares of image attention (cropped for display; full frames in Supp.\ Figs.~\ref{fig:supp_attn_unsink} and \ref{fig:supp_lawn_fullfield}).}
\label{fig:teaser}
\end{figure}

\begin{figure*}[t]
\centering
\includegraphics[width=\textwidth]{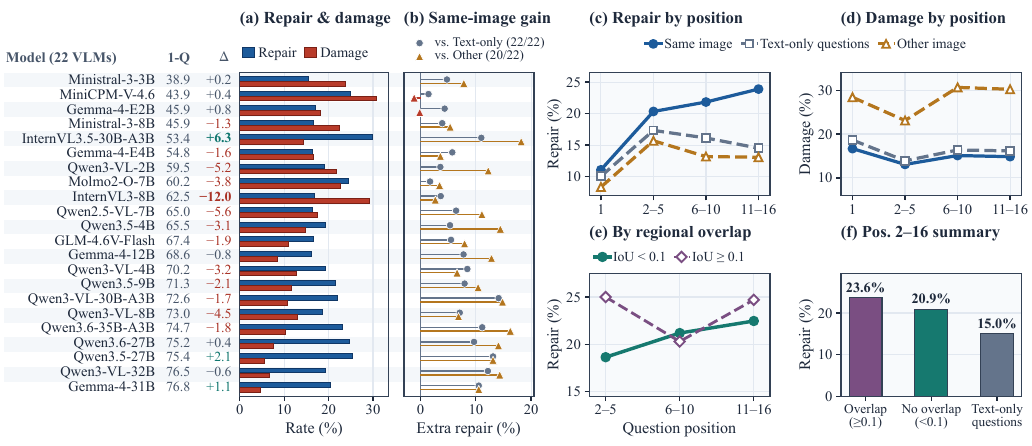}
\caption{\textbf{Repair, damage, and preceding-context effects across $22$ VLMs on DCI} (one shared question order across all $22$ models and panels). (a) Single-question accuracy (\emph{1-Q}), paired co-asking accuracy shift ($\Delta = \text{Co} - \text{1-Q}$, percentage points), and conditional repair (among single-question errors) and damage (among single-question correct answers). (b) Per-model same-image repair gains over text-only questions and other-image questions at Positions~2--16. (c,d) Pooled repair and damage across positions under the three preceding contexts. (e) Pooled same-image repair on the $1{,}402$ regional questions with ($\mathrm{IoU}\ge0.1$) and without ($\mathrm{IoU}<0.1$) prior target-box overlap. (f) Pooled Positions~2--16 repair summary across the same $1{,}402$ regional questions.}
\label{fig:ledger}
\label{fig:context}
\end{figure*}

Practical vision-language applications frequently ask many questions about a single image, such as extracting fields from a form or document~\cite{zheng2021global,tang2024mqma,loem2025joint}, inspecting user interfaces~\cite{wu2023webui}, or querying objects and spatial relations across a busy scene~\cite{sima2024drivelm}.
Rather than issuing a separate query per question, practitioners commonly pack the image and all numbered questions into one prompt before decoding the answers (\emph{co-asking}).
Existing evaluations of co-asking in vision-language models (VLMs), however, report conflicting prompt-level conclusions: joint querying improves field extraction on documents~\cite{loem2025joint}, whereas querying several objects at once on natural scenes increases hallucinations~\cite{chen2024multi}.
Evaluated on Densely Captioned Images (DCI)~\cite{urbanek2024picture} across $22$ open-weight VLMs from seven model families ($1\text{B}$ to $35\text{B}$ parameters)~\cite{bai2025qwen25vl,bai2025qwen3vl,gemmateam2026gemma4,liu2026ministral3,zhu2025internvl3,wang2025internvl35,hong2025glmv,clark2026molmo2,yu2025minicpmv45,openbmb2026minicpmv46}, net accuracy shifts over asking alone span $-12.0$ to $+6.3$ percentage points (\cref{fig:ledger}a) with no consistent trend across model scale.

Pairing each question's single-question and co-asked answers (\cref{fig:ledger}, \cref{sec:ledger}) reveals why: across all $22$ models on DCI, co-asking repairs $15.4\%$--$29.8\%$ of single-question errors regardless of model size or net shift, while the share of single-question correct answers turned wrong drops from $30.8\%$ on weaker models to $4.7\%$ on stronger ones, so the balance between steady repair and falling damage sets each model's net shift.
Before any answer is decoded, reading earlier questions about the same scene raises a later question's attention on its target, relative to same-size random boxes, in all $14$ models supporting custom attention masks (\cref{fig:teaser}, \cref{tab:target_lead}).

Controlled context swaps and causal interventions on DCI trace how this help arises (\cref{sec:repair}).
As more same-image questions precede a query in the $16$-question DCI prompt, repair rises from $11.1\%$ on the first question to $23.9\%$ on Questions~11--16, and it rises even on targets that no earlier question asked about (\cref{fig:ledger}c,e); with length-matched text-only or other-image questions in front, repair instead peaks at Questions~2--5 and then declines.
Earlier questions contribute by reading the image themselves: blocking their attention to the image while keeping their text visible lowers both target localization and accuracy in all $14$ mask-capable models (\cref{tab:knockout_models}).
Blocking one layer window at a time places this reading in the early-to-middle layers, followed $3$--$11$ layers later by sharper target attention on the current question (\cref{fig:cross_question_knockout}).

While cross-question help is established during shared prefill, part of the co-asking damage arises later during answer decoding (\cref{sec:method}, \cref{tab:method_stages}).
Question~1 makes this separation directly visible: when encoding image $V$, instruction $I$, questions $Q_{1..K}$, and answers $A_{1..K}$ as $[V, I, Q_1, \dots, Q_K, A_1, \dots, A_K]$, $Q_1$ precedes $Q_{2..K}$ so under causal masking its prefill states depend only on $V$, $I$, and $Q_1$ itself, yet $A_1$ is decoded \emph{after} $Q_K$ while attending across $Q_{2..K}$; masking $Q_{2..K}$ during decoding cuts Question~1's damaged answers across the $14$ models from $127$ to $61$ ($N = 1{,}384$).
Motivated by this stage asymmetry, we introduce \emph{Read Together, Answer Apart} (\cref{fig:method}), a training-free decoding mask that encodes all questions together in one causal prefill pass and isolates each answer from other stems while re-indexing its position IDs right after its own question~\cite{glavas2026ippd}.
It improves co-asking accuracy on $12$ of $14$ models on DCI (by up to $+4.71$ percentage points; \cref{tab:main_dci}, \cref{fig:method_gain}) and on $10$, $8$, and $5$ of the $14$ models on WebUI~\cite{wu2023webui}, FinTabNet~\cite{zheng2021global}, and GQA~\cite{hudson2019gqa} (by up to $+11.4$, $+5.3$, and $+1.1$ points; \cref{tab:cross_bench_summary}).
Our main contributions are:
\begin{itemize}[leftmargin=*,nosep]
\item \textbf{Question-level decomposition of co-asking.} Pairing individual questions across $22$ VLMs on DCI and $14$ models across WebUI, FinTabNet, and GQA separates consistent cross-question repair ($15.4\%$--$29.8\%$ on DCI) from capability-dependent answer damage.
\item \textbf{Two-step prefill mechanism of cross-question grounding.} Across all $14$ models, earlier same-image questions extract visual context into their tokens in the early-to-middle layers and raise later questions' target attention ratio (adapted from~\cite{zhang2025mllms}) to $\times 1.01$--$\times 1.55$ of asking alone, while repair accumulates even across non-overlapping scene regions.
\item \textbf{Shared reading with isolated answer decoding.} We introduce \emph{Read Together, Answer Apart}, a training-free single-prefill decoding mask that preserves cross-question visual reading while cutting answer-time interference.
\end{itemize}

\section{Related Work}
\label{sec:related}

\paragraph{Multi-question and multi-task prompting.}
In language models, batching several queries into one prompt reduces context overhead~\cite{cheng2023batch,lin2024batchprompt}, and stronger models often gain on composite text tasks~\cite{son2024multi}, though long prompts remain sensitive to query position and surrounding context~\cite{shi2023large,liu2024lost,yue2025efficient}.
In VLMs, beyond sequential visual dialogue~\cite{das2017visual}, multi-question prompts yield conflicting aggregate results: joint querying improves field extraction on document forms~\cite{loem2025joint}, whereas querying five boxed objects at once in ROPE increases hallucinations~\cite{chen2024multi}.
Single-query methods instead decompose one question into sub-questions~\cite{khan2023exploring,zheng2023ddcot} or re-read the prompt~\cite{xu2024rereading,abhinandan2026ask}, while recent analyses note that extra text shifts decision margins~\cite{wang2026when} and flips individual predictions beneath flat aggregate accuracy~\cite{zhang2026illusion}.
We pair each question between single-question and co-asked VLM prompts to separate cross-question repair from damage, and compare same-image questions against length-matched text-only and other-image questions to isolate scene-specific grounding (\cref{sec:ledger,sec:repair}).

\paragraph{Shared image cache and split decoding.}
When multiple queries share an image or text prefix, inference systems reuse key--value caches across branches~\cite{kwon2023efficient,zheng2024sglang,juravsky2024hydragen}, share a live cache across concurrent workers~\cite{rodionov2025hogwild}, or answer follow-up questions from compressed query tokens~\cite{kaduri2025whats}.
Block-attention masks likewise isolate parallel context chunks encoded for a single query~\cite{ratner2023parallel,ma2025block,acharya2025star}.
To answer multiple questions over one shared context without cross-question interaction, Yu and Yao~\cite{yu2026visual} decode each question in an isolated branch after one visual prefill, and Intra-Prompt Parallel Decoding (IPPD)~\cite{glavas2026ippd} packs all questions into one prompt, masks them from one another in both prefill and decoding, and assigns virtual position IDs so that each answer follows its own question.
Both designs aim to reproduce independent single-question outputs; by contrast, we keep causal attention across questions during prefill so later questions benefit from earlier visual reading, and isolate other stems only during answer decoding (\cref{sec:method}).

\paragraph{Visual information flow and localization inside VLMs.}
Mechanistic studies on single image--question pairs examine how visual features transfer to internal representations and generated answers~\cite{basu2024understanding,neo2025towards,jiang2025devils,osaka2026what}, showing that visual information flows from image tokens into question tokens in the early-to-middle layers before propagating to the answer~\cite{kaduri2025whats,zhang2025cross,salazar2026pathways}.
On an isolated question, although attention already favors target regions above chance~\cite{zhang2025mllms}, a large share is absorbed by background sink tokens~\cite{darcet2024vision,xiao2024efficient,kang2025see}, motivating training-free attention calibration and cropping within a single query~\cite{huang2024opera,liu2024paying,zhang2025mllms,li2026cmac}.
Whereas Kaduri et al.~\cite{kaduri2025whats} trace the single-question pathway from the image into that question's own tokens, we show that earlier questions about the same scene pass visual context through their question tokens during prefill to guide later questions onto their targets (\cref{sec:repair}).

\section{Question-Level Analysis of Co-Asking}
\label{sec:ledger}

\begin{figure*}[t]
\centering
\includegraphics[width=\textwidth]{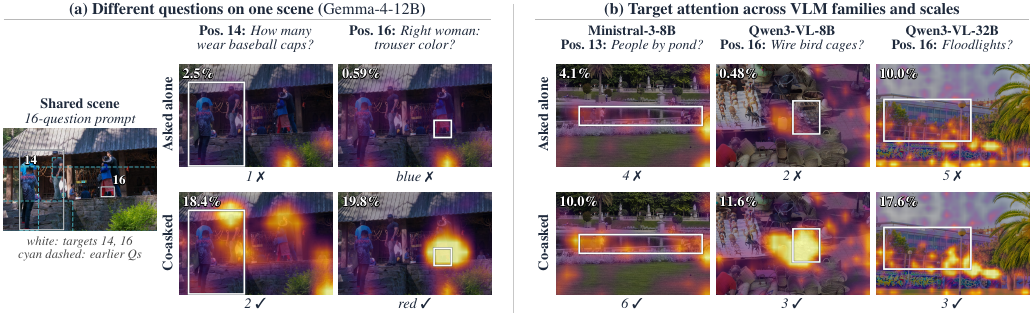}
\caption{\textbf{Co-asking increases prefill attention on the queried target across scenes and VLM families.} (a) On one shared temple scene (Gemma-4-12B, layers $17, 23, 29$ of $48$), asking different questions after earlier scene questions increases final-token image attention on each queried target (white boxes; cyan dashed boxes mark earlier questions). (b) Middle-layer patch attention at the final question token on repaired questions across Ministral-3-8B (layers $12$--$22$ of $34$), Qwen3-VL-8B ($12$--$23$ of $36$), and Qwen3-VL-32B ($35$--$51$ of $64$). Each pair shares a display ceiling; percentages report unsmoothed in-box shares of image attention (Supp.\ Sec.~\ref{sec:supp_ledger}).}
\label{fig:attn_heatmap}
\end{figure*}

\subsection{Experimental Setup}
\label{sec:setup}

We evaluate $22$ open-weight VLMs from seven model families ($1\text{B}$--$35\text{B}$ parameters)~\cite{bai2025qwen25vl,bai2025qwen3vl,gemmateam2026gemma4,liu2026ministral3,zhu2025internvl3,wang2025internvl35,hong2025glmv,clark2026molmo2,yu2025minicpmv45,openbmb2026minicpmv46} under greedy decoding with thinking disabled.
Our primary benchmark is Densely Captioned Images (DCI)~\cite{urbanek2024picture}, comprising $100$ SA-1B~\cite{kirillov2023segment} scenes and $1{,}597$ short-answer questions built from DCI region annotations and verified by hand ($1{,}402$ regional with target boxes and $195$ whole-image, up to $16$ per image), alongside WebUI~\cite{wu2023webui} ($238$ pages, $10$ questions/page), FinTabNet~\cite{zheng2021global} ($300$ tables, $20$ questions/table), and GQA~\cite{hudson2019gqa} ($700$ scenes, $5$ questions/scene).
In a co-asked sequence $[V, I, Q_1, \dots, Q_K, A_1, \dots, A_K]$, image tokens $V$ and instruction $I$ precede all numbered question stems during left-to-right causal prefill, and the model decodes numbered answers one per line, graded by exact match and a frozen Qwen3.8-27B judge while excluding indeterminate judge outputs ($<0.5\%$; prompts, licenses, and examples in Supp.\ Sec.~\ref{sec:supp_setup} and Fig.~\ref{fig:supp_benchmark_examples}).
Pairing each question's single-question answer (asked as ``1.'') with its co-asked answer defines \emph{repair rate} (single-question errors answered right) and \emph{damage rate} (single-question correct answers turned wrong).
Main-paper DCI analyses use one shared question order (three random question orders in Supp.\ Tab.~\ref{tab:supp_dci_14models}).
Behavioral analyses run standard one-pass co-asking across all $22$ models.
In Sections~4--5, where custom 4D attention masks intervene on one question $Q_k$ at a time across the $14$ mask-capable models, changing an earlier answer $A_{<k}$ would alter the text prefix seen by later questions; every masked run and its co-asking baseline therefore decode one answer at a time with earlier answers $A_{<k}$ fixed to the model's co-asking output, so any two compared runs differ only in their attention mask.
Accuracy comparisons among asking alone, co-asking, our method, and blocking use the $1{,}556$--$1{,}594$ of $1{,}597$ questions per model graded as right or wrong across all four settings ($22{,}181$ model--question pairs: $1{,}384$ at Position~1 plus $20{,}797$ at Positions~2--16 where earlier questions exist), while attention-ratio readouts cover the $1{,}356$ Positions~2--16 questions admitting same-size random reference boxes (Supp.\ Sec.~\ref{sec:supp_ledger}).

\begin{table}[t]
\centering
\caption{\textbf{Target attention ratio and earlier-image block across $14$ models on DCI} (Positions~2--16, one shared question order). \emph{Attention ratio / alone}: layer-averaged target-over-random attention ratio (adapted from~\cite{zhang2025mllms}) divided by asking alone ($1.000$) at the final question token under co-asking (Co-ask), after text-only questions (Text-only), and under earlier-image blocking (Blocked, $V \nrightarrow Q_{<k}$; unnormalized values in Supp.\ Tab.~\ref{tab:supp_u4_14models}). $\Delta\text{Acc.}$: accuracy change (\%) from blocking $V \to Q_{<k}$ relative to the unblocked isolated-decoding baseline ($N = 20{,}797$ Positions~2--16 pairs; \cref{sec:repair_knockout}).}
\label{tab:target_lead}\label{tab:knockout_models}
\footnotesize\renewcommand{\arraystretch}{0.89}\setlength{\tabcolsep}{0pt}
\begin{tabular*}{\columnwidth}{@{\extracolsep{\fill}}lrrrr@{}}
\toprule
& \multicolumn{3}{c}{Attention ratio (vs.\ alone $=1.0$)} & $\Delta$Acc. \\
\cmidrule(lr){2-4}
Model & Co-ask & Text-only & Blocked & (\%) \\
\midrule
\raisebox{-0.22ex}{\includegraphics[height=1.65ex]{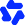}}\hspace{3.2pt}Qwen3-VL-2B & 1.173 & 0.939 & 1.051 & $-2.95$ \\
\raisebox{-0.22ex}{\includegraphics[height=1.65ex]{fig/logos/qwen.pdf}}\hspace{3.2pt}Qwen3-VL-4B & 1.081 & 0.967 & 0.969 & $-2.15$ \\
\raisebox{-0.22ex}{\includegraphics[height=1.65ex]{fig/logos/qwen.pdf}}\hspace{3.2pt}Qwen3-VL-8B & 1.073 & 0.965 & 0.976 & $-3.62$ \\
\raisebox{-0.22ex}{\includegraphics[height=1.65ex]{fig/logos/qwen.pdf}}\hspace{3.2pt}Qwen3-VL-30B-A3B & 1.115 & 0.908 & 0.962 & $-1.94$ \\
\raisebox{-0.22ex}{\includegraphics[height=1.65ex]{fig/logos/qwen.pdf}}\hspace{3.2pt}Qwen3-VL-32B & 1.106 & 0.972 & 1.005 & $-4.05$ \\
\raisebox{-0.22ex}{\includegraphics[height=1.65ex]{fig/logos/qwen.pdf}}\hspace{3.2pt}Qwen2.5-VL-7B & 1.234 & 1.062 & 1.092 & $-1.74$ \\
\raisebox{-0.22ex}{\includegraphics[height=1.65ex]{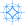}}\hspace{3.2pt}Gemma-4-E2B & 1.554 & 1.355 & 1.534 & $-0.74$ \\
\raisebox{-0.22ex}{\includegraphics[height=1.65ex]{fig/logos/gemma.pdf}}\hspace{3.2pt}Gemma-4-E4B & 1.509 & 1.373 & 1.461 & $-0.07$ \\
\raisebox{-0.22ex}{\includegraphics[height=1.65ex]{fig/logos/gemma.pdf}}\hspace{3.2pt}Gemma-4-12B & 1.277 & 1.185 & 1.224 & $-0.80$ \\
\raisebox{-0.22ex}{\includegraphics[height=1.65ex]{fig/logos/gemma.pdf}}\hspace{3.2pt}Gemma-4-31B & 1.340 & 1.269 & 1.286 & $-2.48$ \\
\raisebox{-0.22ex}{\includegraphics[height=1.65ex]{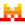}}\hspace{3.2pt}Ministral-3-3B & 1.012 & 0.964 & 0.973 & $-0.62$ \\
\raisebox{-0.22ex}{\includegraphics[height=1.65ex]{fig/logos/mistral.pdf}}\hspace{3.2pt}Ministral-3-8B & 1.077 & 0.972 & 1.018 & $-2.66$ \\
\raisebox{-0.22ex}{\includegraphics[height=1.65ex]{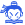}}\hspace{3.2pt}InternVL3-8B & 1.149 & 0.995 & 1.066 & $-0.47$ \\
\raisebox{-0.22ex}{\includegraphics[height=1.65ex]{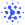}}\hspace{3.2pt}GLM-4.6V-Flash & 1.154 & 0.989 & 1.069 & $-0.54$ \\
\midrule
Pooled (Pos.\ 2--16) & --- & --- & --- & $-1.77$ \\
\bottomrule
\end{tabular*}
\end{table}

\subsection{Repair, Damage, and Target Attention}
\label{sec:ledger_shifts}
\label{sec:ledger_attention}

\textbf{Opposing question-level flips underlie erratic prompt-level totals.}
Ordering all $22$ VLMs by single-question accuracy on DCI (\cref{fig:ledger}a; Supp.\ Tabs.~\ref{tab:supp_ability} and \ref{tab:supp_22models_dci}) shows no consistent link between model capability and net co-asking shift ($\Delta = \text{Co} - \text{1-Q}$, from $-12.0$ points on InternVL3-8B to $+6.3$ on InternVL3.5-30B-A3B): the weakest models barely move in aggregate (Ministral-3-3B $+0.2$, MiniCPM-V-4.6 $+0.4$), whereas mid-ranked models such as Molmo2-O-7B and Qwen3.5-4B drop by $3.8$ and $3.1$ points.
Pairing predictions per question explains why (\cref{fig:ledger}a).
Every one of the $22$ models repairs $15.4\%$--$29.8\%$ of its single-question errors under co-asking on DCI, with consistent one-pass co-asking repair across all $14$ evaluated models on WebUI ($12.9\%$--$25.9\%$), FinTabNet ($14.7\%$--$46.8\%$), and GQA ($7.8\%$--$19.0\%$; Supp.\ Tab.~\ref{tab:supp_cross_native_repair}).
Damage moves sharply in the opposite direction across capability: the share of single-question correct answers turned wrong falls from $23.8\%$--$30.8\%$ on Ministral-3-3B and MiniCPM-V-4.6 to $4.7\%$--$6.6\%$ on Gemma-4-31B, Qwen3.5-27B, and Qwen3-VL-32B, so the share of all answers that co-asking turns wrong largely decides whether a model's net score rises or falls.

\textbf{Co-asking raises question-to-image attention on the queried target.}
\begin{figure*}[t]
\centering
\includegraphics[width=\textwidth]{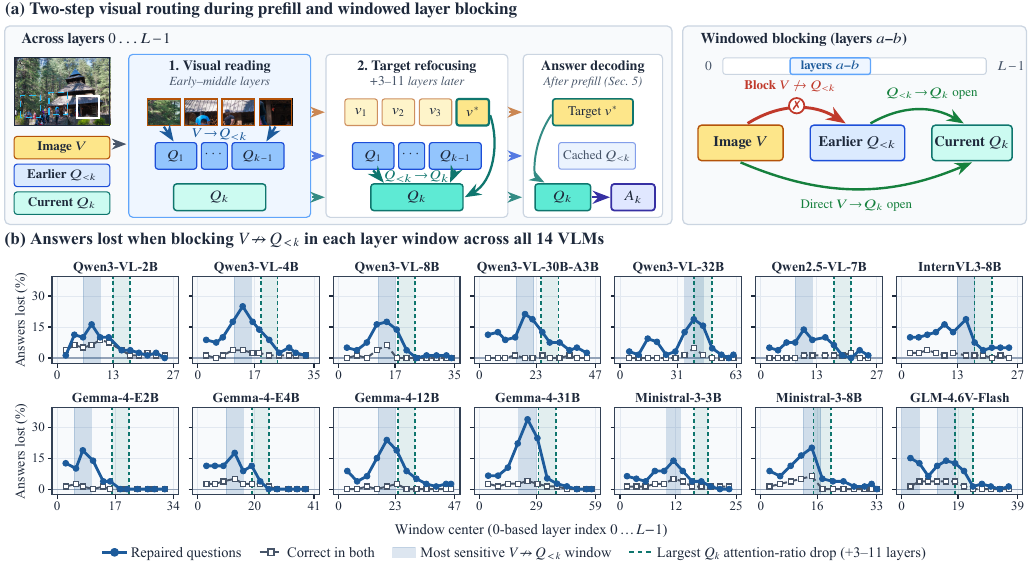}
\caption{\textbf{Two-step cross-question visual routing across all $14$ models.} (a) Two-step prefill routing across layers $0 \dots L-1$ (1: earlier questions $Q_{<k}$ read image patches; 2: current question $Q_k$ reads $Q_{<k}$ and shifts attention from background patches $v_1 \dots v_3$ onto its queried target patch $v^*$ before answer decoding) and windowed blocking of earlier stems from attending to the image ($V \nrightarrow Q_{<k}$) within layers $a$--$b$ while leaving $Q_{<k} \to Q_k$ and direct $V \to Q_k$ open. (b) Percentage of answers lost per layer window across all $14$ models on repaired questions versus questions answered right in both settings (Positions~2--16, one question order). Blue shading marks the most sensitive $V \nrightarrow Q_{<k}$ windows (both peaks shaded on bimodal GLM-4.6V-Flash); dashed teal spans mark layers with the largest drop in $Q_k$ target attention ratio.}
\label{fig:cross_question_knockout}
\end{figure*}
Because image tokens precede all questions under causal masking, adding questions leaves the image token representations unchanged; any cross-question change therefore comes from how later question and answer tokens attend to the prompt.
Inspecting prefill question-to-image attention before any answer is decoded (\cref{fig:attn_heatmap}) reveals a consistent visual shift: on questions missed alone, middle-layer attention largely scatters over uninformative background patches~\cite{darcet2024vision,kang2025see}, whereas asking after earlier scene questions increases attention on each question's own target both across questions on the same scene (\cref{fig:attn_heatmap}a: $2.5\% \to 18.4\%$ at Position~14 and $0.59\% \to 19.8\%$ at Position~16) and across VLM families (\cref{fig:attn_heatmap}b).
To quantify target localization across all $14$ models while controlling for target-box area, we adapt the attention ratio of Zhang et al.~\cite{zhang2025mllms}: at the final question token across Positions~2--16, we divide each question's layer-averaged image attention on the ground-truth target box by its mean over same-size non-overlapping random boxes, without normalizing by a generic-description prompt ($1.0$ corresponds to chance, with raw single-question ratios already spanning $1.75$--$4.01$; Supp.\ Tab.~\ref{tab:supp_u4_14models}).
Dividing each condition's ratio by asking alone ($=1.000$ in \cref{tab:target_lead}) shows that co-asking raises target focus to $\times 1.01$--$\times 1.55$ in all $14$ models, even though earlier questions usually query other parts of the scene.

\section{Cross-Question Visual Grounding}
\label{sec:repair}

\subsection{Preceding Context and Regional Overlap}
\label{sec:repair_same_image}
\label{sec:repair_whole_image}
\label{sec:repair_scaling}

\textbf{Scene-specific context accumulates across positions.}
To test whether cross-question repair comes from scene-specific visual context rather than prompt length or numbered question syntax, we compare three preceding contexts at matched length and position while always viewing the target image (\cref{fig:ledger}b,c, \cref{tab:target_lead}; examples in Supp.\ Fig.~\ref{fig:supp_benchmark_examples}b and per-model rates in Supp.\ Tab.~\ref{tab:supp_22models_dci}): earlier questions about the \emph{same image}, synthetic image-free \emph{text-only questions} (arithmetic and word tasks), and questions written for \emph{another, unseen image} (evaluated with a shared unanswerable-question clause; Supp.\ Sec.~\ref{sec:supp_repair}).
Pooled across all $22$ models (\cref{fig:ledger}c), Position~1 already shows small shifts ($11.1\%$, $10.0\%$, and $8.3\%$) because under one-pass co-asking its answer $A_1$ is decoded after trailing stems $Q_{2..16}$ (\cref{sec:method}); once earlier questions precede the query during prefill, repair after same-image questions climbs steadily to $20.3\%$, $21.8\%$, and $23.9\%$ across Positions~2--5, 6--10, and 11--16 ($22.3\%$ across Positions~2--16).
By contrast, repair after text-only questions ($15.7\%$ across Positions~2--16) or other-image questions ($13.7\%$) peaks at Positions~2--5 and then declines.
Across Positions~2--16, same-image questions yield higher repair than text-only questions on all $22$ models ($+1.4$ to $+14.1$ points), beat other-image questions on $20$ of $22$ (\cref{fig:ledger}b), and produce a higher target attention ratio than text-only questions in all $14$ models (\cref{tab:target_lead}).

\textbf{Cross-question help extends across non-overlapping regions.}
Splitting the $1{,}402$ regional DCI questions at Positions~2--16 by whether the current target box overlaps any earlier question's box ($\text{IoU} \ge 0.1$ vs.\ $\text{IoU} < 0.1$; \cref{fig:ledger}e,f and Supp.\ Tab.~\ref{tab:supp_nonoverlap}), overlapping targets start higher at Positions~2--5 ($25.0\%$ vs.\ $18.6\%$ pooled repair across all $22$ models).
As more questions query other parts of the scene, however, repair on non-overlapping targets ($\text{IoU} < 0.1$) climbs steadily from $18.6\%$ at Positions~2--5 to $21.2\%$ at Positions~6--10 and $22.5\%$ at Positions~11--16 (\cref{fig:ledger}e), exceeding text-only questions on the same regional set across Positions~2--16 ($20.9\%$ vs.\ $15.0\%$, \cref{fig:ledger}f).
Asking about diverse parts of an image thus progressively builds scene-wide context that helps later questions locate new targets without querying the same object twice.

\subsection{Blocking Earlier Image Attention}
\label{sec:repair_knockout}

Because co-asking also exposes $Q_k$ to the text of earlier stems $Q_{<k}$, comparing co-asking with single-question prompting alone cannot separate visual reading from textual clues across stems.
To isolate visual reading during prefill, for each target question $Q_k$ ($k \ge 2$) we keep all earlier stems $Q_{<k}$ in the prompt and leave both $Q_{<k} \to Q_k$ and direct $V \to Q_k$ open, blocking only earlier stems from attending to the image ($V \nrightarrow Q_{<k}$; \cref{fig:cross_question_knockout}a).
Unlike asking alone, $Q_k$ still reads the full text of $Q_{<k}$ during prefill, so any lexical or format cues from earlier stems remain available; both the blocked run and its unblocked baseline also mask non-target stems during answer decoding with fixed prior answers $A_{<k}$ (\cref{sec:setup,sec:method}) so answer-time stem interference cannot confound the comparison.

Across the $20{,}797$ Positions~2--16 pairs, blocking $V \to Q_{<k}$ lowers accuracy in all $14$ models ($-1.77$ percentage points pooled on DCI, erasing $24.6\%$ of repaired answers pooled and $19.3\%$--$36.9\%$ per model; \cref{tab:knockout_models}, Supp.\ Sec.~\ref{sec:supp_repair}).
Losses deepen as more earlier questions accumulate, growing from $-1.06$ points at Positions~2--5 to $-1.41$ at Positions~6--10 and $-2.56$ at Positions~11--16 (Supp.\ Sec.~\ref{sec:supp_repair}).
Target localization drops in parallel: in every model, blocking $V \to Q_{<k}$ reduces the current question's target attention ratio from $\times 1.01$--$\times 1.55$ under co-asking to $\times 0.96$--$\times 1.53$ (\cref{tab:knockout_models}).
Earlier questions thus contribute beyond textual clues by attending to the image during prefill, helping later questions locate their targets.

\subsection{Layer-Window Analysis}
\label{sec:repair_layers}

To locate which layers carry out this cross-question reading, we follow Zhang et al.~\cite{zhang2025cross} and block $V \to Q_{<k}$ inside one contiguous layer window at a time ($4$--$11$ layers, roughly one-sixth of model depth across $12$ overlapping windows per model, $13$ on $64$-layer Qwen3-VL-32B; Supp.\ Tab.~\ref{tab:supp_window_sweep}).
In $36$-layer Qwen3-VL-8B, for instance, each pass blocks $V \to Q_{<k}$ inside a single span (layers $0$--$5$, $3$--$8$, $\dots$, $30$--$35$, plus tail $33$--$35$) while leaving all other layers untouched, and we compare the share of answers lost across up to $80$ sampled questions per group: questions repaired over asking alone under the unblocked baseline versus questions answered right in both settings (\cref{fig:cross_question_knockout}b; Supp.\ Tab.~\ref{tab:supp_window_sweep}).
Across all $14$ models, losses on repaired questions concentrate in the early-to-middle layers: the most sensitive window centers at $30\%$--$64\%$ of depth across $12$ models---shifting earlier only in $35$-layer Gemma-4-E2B (layers $5$--$10$) and bimodal $40$-layer GLM-4.6V-Flash (layers $0$--$6$ and $12$--$18$)---whereas blocking deep layers has almost no effect ($0.0\%$--$5.0\%$).
Blocking $V \to Q_{<k}$ in that single peak window erases $13.8\%$--$33.8\%$ of repaired answers across the $14$ models, versus only $0.0\%$--$6.3\%$ on questions answered right in both settings (Supp.\ Tab.~\ref{tab:supp_window_sweep}), confirming that the window block selectively removes cross-question help rather than disrupting single-question perception.

To see when that visual signal reaches the current question $Q_k$, we track $Q_k$'s layer-by-layer target attention ratio when $V \to Q_{<k}$ is blocked across all layers relative to unblocked co-asking: the window where $Q_k$'s target attention ratio drops most (\cref{fig:cross_question_knockout}b, dashed spans; Supp.\ Tab.~\ref{tab:supp_window_sweep}) begins $3$--$11$ layers \emph{after} the most sensitive reading window of $Q_{<k}$ across all $14$ models---for example, layers $40$--$50$ after $35$--$45$ in $64$-layer Qwen3-VL-32B, layers $24$--$31$ after $16$--$23$ in $48$-layer Gemma-4-12B, and layers $18$--$24$ after the middle peak $12$--$18$ in GLM-4.6V-Flash.
Together, these interventions trace a two-step prefill pathway (\cref{fig:cross_question_knockout}a) extending single-question visual information flow~\cite{kaduri2025whats,zhang2025cross}: earlier questions $Q_{<k}$ first read visual information from $V$ into their token states in the early-to-middle layers, and over the next few layers, $Q_k$ attends to $Q_{<k}$ and focuses onto its own target before decoding begins.

\begin{figure*}[t]
\centering
\includegraphics[width=\textwidth]{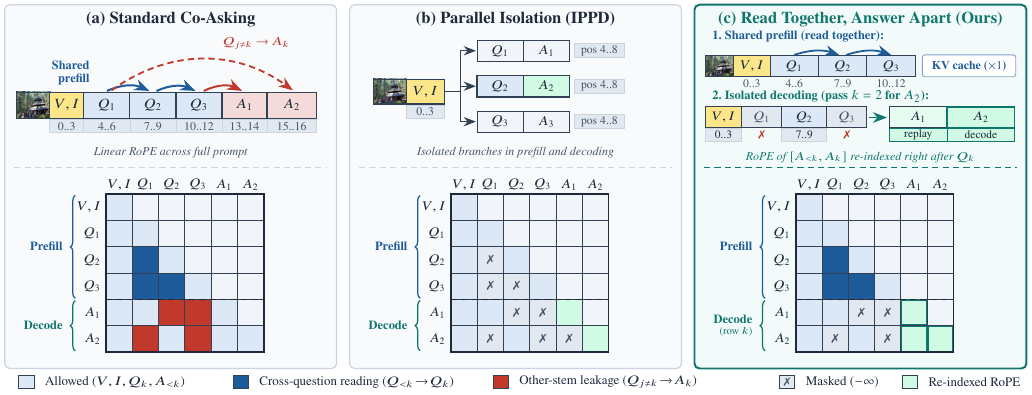}
\caption{\textbf{Shared question prefill and isolated answer decoding.} (a) Standard co-asking encodes $[V, I, Q_1, \dots, Q_K]$ under left-to-right causal masking during prefill (blue cells), then exposes each decoded answer $A_k$ to all other question stems $Q_{j \ne k}$ (red cells). (b) IPPD~\cite{glavas2026ippd} masks cross-question attention in both prefill and decoding. (c) \emph{Read Together, Answer Apart} (Ours) keeps one shared causal prefill pass so later questions read earlier ones, then isolates each decoded answer $A_k$ from non-target stems $Q_{j \ne k}$ while re-indexing $[A_{<k}, A_k]$ consecutively after $Q_k$. Each \emph{Decode} row shows that answer's own mask $M_k$ (\cref{eq:mask}; line-break handling per tokenizer in Supp.\ Sec.~\ref{sec:supp_method}).}
\label{fig:method}
\end{figure*}

\section{Read Together, Answer Apart}
\label{sec:method}
\label{sec:damage}

\subsection{Answer-Time Interference and Method}
\label{sec:damage_load}
\label{sec:damage_answering}
\label{sec:method_rule}

\begin{table}[t]
\centering
\caption{\textbf{Repair and damage counts by question position on DCI} (pooled across all $14$ models under one question order, $N = 22{,}181$: $1{,}384$ at Position~1 $+$ $20{,}797$ at Positions~2--16). \emph{Repair}: wrong alone $\to$ right; \emph{Damage}: right alone $\to$ wrong.}
\label{tab:method_stages}\label{tab:paired_changes}
\footnotesize
\setlength{\tabcolsep}{0pt}
\renewcommand{\arraystretch}{0.92}
\begin{tabular*}{\columnwidth}{@{\extracolsep{\fill}}lrrrrr@{}}
\toprule
Position & $N$ & \multicolumn{2}{c}{Repair} & \multicolumn{2}{c}{Damage} \\
\cmidrule(lr){3-4}\cmidrule(lr){5-6}
 & & Co-asking & Ours & Co-asking & Ours \\
\midrule
1 & 1,384 & 53 & 24 & 127 & 61 \\
2--16 & 20,797 & 1,408 & 1,494 & 1,901 & 1,625 \\
\midrule
All (1--16) & 22,181 & 1,461 & 1,518 & 2,028 & 1,686 \\
\bottomrule
\end{tabular*}
\end{table}

Packing multiple questions into one prompt adds context load: pooled across all $22$ models at Positions~2--16 on DCI (\cref{fig:ledger}d; per-model rates in Supp.\ Tab.~\ref{tab:supp_22models_dci}), preceding a question with text-only questions turns $15.6\%$ of single-question correct answers wrong, and questions from another image turn $28.3\%$ wrong, compared with $14.4\%$ under same-image questions.
Removing earlier questions from the prompt would avoid this load, but would also forfeit the visual grounding they build during prefill (\cref{sec:repair}).
Under left-to-right causal masking in $[V, I, Q_1, \dots, Q_K, A_1, \dots, A_K]$, however, each question $Q_k$ reads only earlier stems $Q_{<k}$ during prefill, whereas each answer $A_k$ is decoded \emph{after} $Q_K$ with all other question stems $Q_{j \ne k}$ still visible (\cref{fig:method}a).

Motivated by this stage asymmetry, \emph{Read Together, Answer Apart} (\cref{fig:method}c) keeps shared question prefill intact and isolates only answer decoding, without training or prompt changes:

\textbf{Shared reading (prefill).}
We encode $[V, I, Q_1, \dots, Q_K]$ once under standard left-to-right causal attention and cache its keys and values across all layers, so every question $Q_k$ reads $V$, $I$, and all earlier stems $Q_{<k}$ at its natural prompt position.

\textbf{Isolated answering (decoding).}
Reusing the cached prefix, we decode each answer $A_k$ ($k = 1, \dots, K$) with the decoder attending to $V$, $I$, its own question $Q_k$, and the numbered prior-answer prefix $A_{<k}$ (\cref{sec:setup}), while masking out all other question stems $Q_{j \ne k}$ (and masking later stems $Q_{j > k}$ from $A_{<k}$) across all layers:
{\setlength{\abovedisplayskip}{3pt}\setlength{\belowdisplayskip}{3pt}\small
\begin{equation}
    M_k(u, v) =
    \begin{cases}
        -\infty, & v > u,\text{ or } u \in A_k,\; v \in \textstyle\bigcup_{j \ne k} Q_j \\[-2pt]
                 & \text{or } u \in A_{<k},\; v \in \textstyle\bigcup_{j > k} Q_j, \\
        0,       & \text{otherwise.}
    \end{cases}
    \label{eq:mask}
\end{equation}}
Masking $Q_{j > k}$ from $A_{<k}$ as well prevents fixed prior answers from relaying later question stems into $A_k$.
Because $K - k$ masked stems sit between $Q_k$ and $A_1$ in the raw prompt, leaving position IDs unchanged would insert an artificial distance gap that widens for earlier questions; following IPPD~\cite{glavas2026ippd}, we re-index the rotary position IDs of $[A_{<k}, A_k]$ consecutively right after $Q_k$ so each answer is decoded adjacent to its own stem (Supp.\ Sec.~\ref{sec:supp_method}).
For Question~2 ($k=2$ in \cref{fig:method}c), $Q_2$'s cached prefill states already carry the visual context read through $Q_1$, while decoding $A_2$ masks both $Q_1$ and $Q_3$ and places $[A_1, A_2]$ directly after $Q_2$.
Unlike IPPD (\cref{fig:method}b), which masks cross-question attention in both prefill and decoding to reproduce independent single-question queries (\emph{Single-Q})---forfeiting cross-question repair---our mask preserves shared prefill reading.

\begin{figure*}[t]
\begin{minipage}[t]{0.485\textwidth}\vspace{0pt}
\centering
{\captionof{table}{\textbf{DCI accuracy (\%) across $14$ models} (one question order, $N = 22{,}181$). Single-Q asks each question alone; Co-Asking decodes per question like Ours without the answer-time mask; $\Delta = \text{Ours} - \text{Co-Asking}$ from unrounded counts (Supp.\ Tab.~\ref{tab:supp_dci_14models}).}\label{tab:main_dci}}
\footnotesize
\renewcommand{\arraystretch}{0.86}
\setlength{\tabcolsep}{0pt}
\begin{tabular*}{\linewidth}{@{\extracolsep{\fill}}lrrrr@{}}
\toprule
Model & Single-Q & Co-Asking & \textbf{Ours} & $\Delta$ \\
\midrule
\raisebox{-0.22ex}{\includegraphics[height=1.65ex]{fig/logos/qwen.pdf}}\hspace{3.2pt}Qwen2.5-VL-7B & 65.0 & 59.8 & \textbf{64.5} & $+4.71$ \\
\raisebox{-0.22ex}{\includegraphics[height=1.65ex]{fig/logos/qwen.pdf}}\hspace{3.2pt}Qwen3-VL-2B & 59.5 & 54.1 & \textbf{58.1} & $+4.03$ \\
\raisebox{-0.22ex}{\includegraphics[height=1.65ex]{fig/logos/internvl.pdf}}\hspace{3.2pt}InternVL3-8B & 62.5 & 49.2 & \textbf{53.0} & $+3.84$ \\
\raisebox{-0.22ex}{\includegraphics[height=1.65ex]{fig/logos/mistral.pdf}}\hspace{3.2pt}Ministral-3-3B & 38.9 & 38.6 & \textbf{42.2} & $+3.60$ \\
\raisebox{-0.22ex}{\includegraphics[height=1.65ex]{fig/logos/mistral.pdf}}\hspace{3.2pt}Ministral-3-8B & 45.9 & 45.0 & \textbf{47.3} & $+2.30$ \\
\raisebox{-0.22ex}{\includegraphics[height=1.65ex]{fig/logos/qwen.pdf}}\hspace{3.2pt}Qwen3-VL-8B & 73.0 & 69.1 & \textbf{71.0} & $+1.88$ \\
\raisebox{-0.22ex}{\includegraphics[height=1.65ex]{fig/logos/qwen.pdf}}\hspace{3.2pt}Qwen3-VL-30B-A3B & 72.6 & 71.4 & \textbf{73.3} & $+1.88$ \\
\raisebox{-0.22ex}{\includegraphics[height=1.65ex]{fig/logos/glm.pdf}}\hspace{3.2pt}GLM-4.6V-Flash & 67.4 & 65.6 & \textbf{67.3} & $+1.64$ \\
\raisebox{-0.22ex}{\includegraphics[height=1.65ex]{fig/logos/gemma.pdf}}\hspace{3.2pt}Gemma-4-12B & 68.6 & 67.6 & \textbf{69.1} & $+1.44$ \\
\raisebox{-0.22ex}{\includegraphics[height=1.65ex]{fig/logos/qwen.pdf}}\hspace{3.2pt}Qwen3-VL-4B & 70.2 & 67.3 & \textbf{68.6} & $+1.26$ \\
\raisebox{-0.22ex}{\includegraphics[height=1.65ex]{fig/logos/qwen.pdf}}\hspace{3.2pt}Qwen3-VL-32B & 76.5 & 76.2 & \textbf{76.7} & $+0.57$ \\
\raisebox{-0.22ex}{\includegraphics[height=1.65ex]{fig/logos/gemma.pdf}}\hspace{3.2pt}Gemma-4-31B & 76.8 & 78.4 & \textbf{78.5} & $+0.06$ \\
\raisebox{-0.22ex}{\includegraphics[height=1.65ex]{fig/logos/gemma.pdf}}\hspace{3.2pt}Gemma-4-E4B & 54.8 & \textbf{53.1} & 52.7 & $-0.38$ \\
\raisebox{-0.22ex}{\includegraphics[height=1.65ex]{fig/logos/gemma.pdf}}\hspace{3.2pt}Gemma-4-E2B & 45.9 & \textbf{46.9} & 45.2 & $-1.65$ \\
\midrule
Pooled ($14$ models) & 62.8 & 60.2 & \textbf{62.0} & $+1.80$ \\
\bottomrule
\end{tabular*}
\end{minipage}\hfill
\begin{minipage}[t]{0.485\textwidth}\vspace{0pt}\centering
\includegraphics[width=\linewidth]{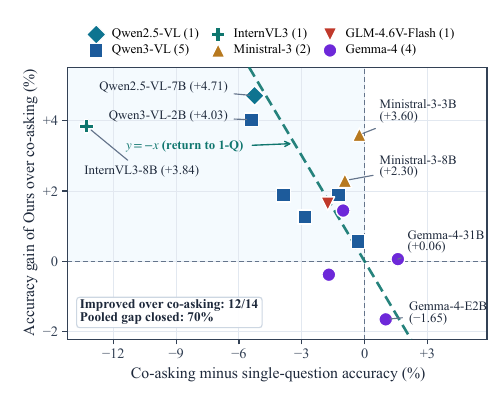}
\caption{\textbf{Gain over co-asking vs.\ co-asking gap across $14$ models on DCI.} Dashed line ($y = -x$): single-question accuracy.}\label{fig:method_gain}
\end{minipage}
\end{figure*}

\subsection{Experimental Results}
\label{sec:experiments_main}

Question~1 separates the two stages directly (\cref{tab:method_stages}).
During prefill, $Q_1$ precedes every other question, so under causal attention its token states depend only on $V$, $I$, and $Q_1$ itself; its co-asked answer can still change because $A_1$ is decoded after $Q_K$ and attends across $Q_{2..K}$.
Masking $Q_{2..K}$ while decoding $A_1$ cuts damaged answers on Question~1 from $127$ to $61$ across the $14$ models; decoding-time repairs (where attending across $Q_{2..K}$ flipped an uncertain answer right) also drop from $53$ to $24$, yielding $+37$ net correct answers.
At Positions~2--16, where each question has already read earlier ones during prefill, our method keeps the repaired answers ($1{,}408 \to 1{,}494$) while cutting damaged answers from $1{,}901$ to $1{,}625$ (\cref{tab:method_stages}).

Across all positions on DCI (\cref{tab:main_dci}; three-order robustness and closed-candidate log-likelihood margins in Supp.\ Tabs.~\ref{tab:supp_dci_14models} and \ref{tab:supp_cand_margin}), \emph{Read Together, Answer Apart} improves accuracy over co-asking on $12$ of the $14$ models ($11$ of the $12$ models hurt by co-asking) by up to $+4.71$ points, led by Qwen2.5-VL-7B ($+4.71$), Qwen3-VL-2B ($+4.03$), InternVL3-8B ($+3.84$), Ministral-3-3B ($+3.60$), and Ministral-3-8B ($+2.30$).
Pooled across all $14$ models, co-asking trails asking alone by $2.6$ points ($60.2\%$ vs.\ $62.8\%$); our method closes $70\%$ of this gap to $62.0\%$ ($+1.80$ points over co-asking; \cref{fig:method_gain}) and matches or exceeds single-question accuracy on $6$ of $14$ models.

\Cref{tab:cross_bench_summary} summarizes all $14$ models on WebUI, FinTabNet, and GQA.
On WebUI, blocking earlier questions from the image ($V \nrightarrow Q_{<k}$) lowers Positions~2--10 accuracy in all $14$ models ($-2.87$ points pooled; $-0.7$ to $-5.1$ per model, Supp.\ Tab.~\ref{tab:supp_cross_dataset}), extending cross-question visual reading beyond natural scenes.
Our method exceeds co-asking on $10$ of $14$ models on WebUI ($67.3\% \to 69.6\%$ pooled, up to $+11.4$ on InternVL3-8B and $+10.8$ on Qwen3-VL-2B), $8$ of $14$ on FinTabNet ($46.6\% \to 46.3\%$ pooled, up to $+5.3$ on InternVL3-8B and $+3.0$ on Qwen2.5-VL-7B), and $5$ of $14$ on $5$-question GQA ($64.4\% \to 64.4\%$ pooled, up to $+1.1$ on InternVL3-8B), where short prompts incur no net co-asking drop ($64.4\%$ vs.\ $64.3\%$) and leave little interference.
Wherever our method trails co-asking, the drop stays below $1.0$ point on all four benchmarks.
The only exceptions are the two smallest Gemma-4 models: Gemma-4-E2B on DCI ($-1.65$), and Gemma-4-E2B and Gemma-4-E4B on $20$-question FinTabNet ($-12.5$ and $-2.5$).
These two models account for FinTabNet's $-0.3$ pooled shift; pooled over the other $12$ models, our method exceeds co-asking by $0.9$ points ($45.9\% \to 46.8\%$).

\begin{table}[t!]
\centering
\caption{\textbf{Cross-benchmark summary across all $14$ models.} \emph{Repair}: single-question errors fixed under one-pass co-asking (range over the $14$ models; Supp.\ Tabs.~\ref{tab:supp_22models_dci} and \ref{tab:supp_cross_native_repair}). \emph{1-Q / Co / Ours}: count-pooled accuracies (\%); \emph{Improved}: models where \emph{Ours} exceeds co-asking, with max gain in points (Supp.\ Tab.~\ref{tab:supp_cross_dataset}).}
\label{tab:cross_bench_summary}
\footnotesize
\renewcommand{\arraystretch}{0.89}
\setlength{\tabcolsep}{0pt}
\begin{tabular*}{\columnwidth}{@{\extracolsep{\fill}}lcrrrr@{}}
\toprule
Benchmark & Repair (\%) & 1-Q & Co & \textbf{Ours} & Improved (max $\Delta$) \\
\midrule
DCI ($16\text{Q}$)       & $15.4\text{--}22.0$ & 62.8 & 60.2 & \textbf{62.0} & $12/14$ ($+4.7$) \\
WebUI ($10\text{Q}$)     & $12.9\text{--}25.9$ & 72.6 & 67.3 & \textbf{69.6} & $10/14$ ($+11.4$) \\
FinTabNet ($20\text{Q}$) & $14.7\text{--}46.8$ & 46.9 & \textbf{46.6} & 46.3 & $8/14$ ($+5.3$) \\
GQA ($5\text{Q}$)        & $7.8\text{--}19.0$  & 64.3 & \textbf{64.4} & \textbf{64.4} & $5/14$ ($+1.1$) \\
\bottomrule
\end{tabular*}
\end{table}

\section{Discussion and Conclusion}
\label{sec:conclusion}

Pairing individual questions across $22$ open-weight VLMs shows that erratic prompt-level co-asking totals reflect two opposing effects: consistent cross-question repair ($15.4\%$--$29.8\%$ on DCI) in every model alongside capability-dependent answer damage.
Across all $14$ models, attention readouts and layer-window interventions trace this repair to a two-step prefill pathway in the early-to-middle layers, where earlier questions read visual context into their tokens and help later questions focus on their targets.

Exploiting this asymmetry, \emph{Read Together, Answer Apart} needs no training: one shared causal prefill and an answer-time mask preserve cross-question repair and cut answer-time interference on $11$ of $12$ models hurt by co-asking.

\textbf{Limitations and outlook.}
Our method requires custom attention masks with key--value caching and leaves prefill context load unchanged; on $20$-question tables, the two smallest Gemma-4 models trail co-asking (\cref{sec:experiments_main}), and understanding why is left for future work.
More broadly, causal attention keeps image tokens independent of later questions, so the cross-question gain comes from how questions read the image.
We hope this finding opens a new direction: using the image-reading behavior that arises in multi-question prompts to improve single-question VQA, for example by distilling it into single-question inference.

{
    \small
    \bibliographystyle{ieeenat_fullname}
    \bibliography{refs}
}

\clearpage
\makeatletter
\setlength{\@fptop}{0pt}\setlength{\@fpsep}{14pt}\setlength{\@fpbot}{0pt plus 1fil}
\setlength{\@dblfptop}{0pt}\setlength{\@dblfpsep}{14pt}\setlength{\@dblfpbot}{0pt plus 1fil}
\makeatother
\maketitlesupplementary
\appendix
\setcounter{table}{0}
\setcounter{figure}{0}
\renewcommand{\thetable}{S\arabic{table}}
\renewcommand{\thefigure}{S\arabic{figure}}
\renewcommand{\theHtable}{S\arabic{table}}
\renewcommand{\theHfigure}{S\arabic{figure}}
\renewcommand{\dbltopfraction}{0.95}
\renewcommand{\topfraction}{0.90}
\renewcommand{\bottomfraction}{0.80}
\renewcommand{\dblfloatpagefraction}{0.75}
\renewcommand{\floatpagefraction}{0.75}
\renewcommand{\textfraction}{0.06}

\noindent
This supplementary material provides additional experimental details and per-model results: setup, prompt templates, grading rules, and asset licenses (\cref{sec:supp_setup}); per-model DCI breakdowns, target-attention ratios, and patch-grid controls (\cref{sec:supp_ledger}); preceding-context controls and layer-windowed blocking (\cref{sec:supp_repair}); answer-time masking and rotary position re-indexing (\cref{sec:supp_method}); question-order robustness and candidate log-likelihood margins (\cref{sec:supp_three_orders}); and per-model evaluations on WebUI, FinTabNet, and GQA (\cref{sec:supp_additional}).

\section{Experimental Setup}
\label{sec:supp_setup}

\begin{figure*}[tp]
\centering
\includegraphics[width=\textwidth]{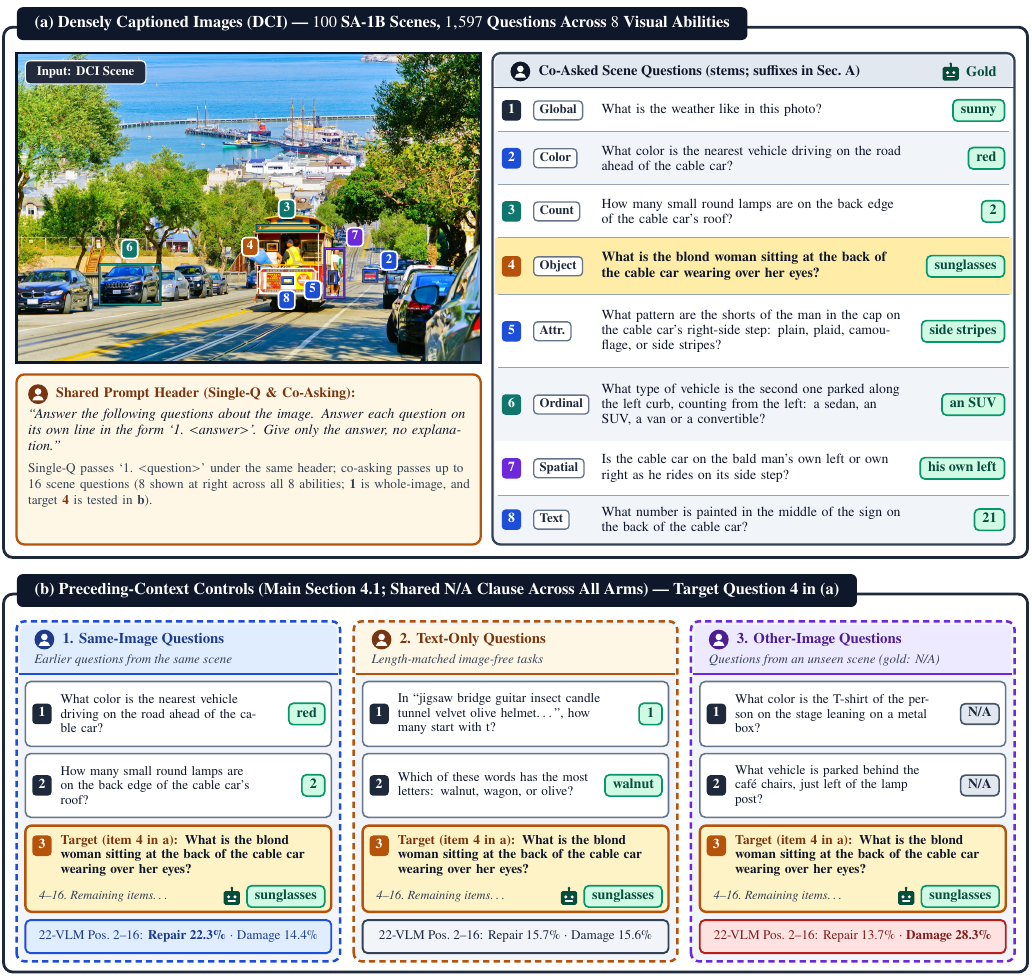}
\caption{\textbf{Representative DCI scene, co-asked questions, and preceding-context controls} (verbatim stems in~a, compactly displayed in~b; format suffixes in \cref{sec:supp_setup}; benchmark examples for WebUI, FinTabNet, and GQA appear in \cref{fig:supp_webui_example,fig:supp_fintab_example,fig:supp_gqa_example}). (a) Representative DCI harbor scene with $8$ of its $16$ co-asked questions spanning all eight visual ability categories under the shared prompt header. (b) Side-by-side comparison of the three preceding-context controls (main Section~4.1 and \cref{sec:supp_repair}) evaluating the same Position-3 target question (\textbf{4} in~a) under same-image, length-matched text-only, and unseen other-image preceding questions with a shared \emph{N/A} clause, alongside 22-VLM pooled Positions~2--16 repair and damage rates.}
\label{fig:supp_benchmark_examples}
\end{figure*}

\paragraph{Benchmarks, prompt templates, and licenses.}
Every prompt places the image tokens first, followed by \emph{``Answer the following questions about the image. Answer each question on its own line in the form `1.\ \textless answer\textgreater'. Give only the answer, no explanation.''} and the numbered question(s) (\cref{fig:supp_benchmark_examples}a, passing a single item `1.\ \textless question\textgreater' in single-question runs and up to $16$ items in co-asked runs).
Format constraints attach only to the question stem they govern: color questions append \emph{``(Answer with exactly one of: red, orange, yellow, green, blue, purple, pink, brown, black, white, gray.)''}, count questions append \emph{``(Answer with a number in digits.)''}, and multiple-choice questions append \emph{``(Answer with one of the listed options.)''}.
On DCI~\cite{urbanek2024picture}, we evaluate $100$ SA-1B~\cite{kirillov2023segment} scenes containing $1{,}597$ short-answer questions constructed from DCI region annotations and verified by the authors across eight visual ability categories ($1{,}402$ regional questions with target boxes and $195$ whole-image questions, including $414$ multiple-choice items; \cref{tab:supp_ability}).
On additional benchmarks (\cref{fig:supp_webui_example,fig:supp_fintab_example,fig:supp_gqa_example}), we evaluate WebUI~\cite{wu2023webui} ($238$ webpage screenshots $\times 10 = 2{,}380$ short-answer questions derived from element bounding-box and hierarchy annotations), FinTabNet~\cite{zheng2021global} ($300$ financial tables $\times 20 = 6{,}000$ cell-lookup, comparison, spatial, and header questions derived from table-structure annotations), and GQA~\cite{hudson2019gqa} ($700$ scenes $\times 5$ entity-disjoint questions $= 3{,}500$ unique questions rotated across $10$ balanced question orders in which each question appears twice at each position $1\text{--}5$, totaling $35{,}000$ observations per model).
DCI annotations are used under CC BY-NC~4.0, SA-1B images under the SA-1B Research License, WebUI code and annotations under the MIT License (with webpage screenshots used solely for academic evaluation), FinTabNet under CDLA-Permissive-1.0, and GQA annotations under CC BY~4.0 with Visual Genome images; all $22$ open-weight VLMs (\cref{tab:supp_ability}, evaluated in BF16 with greedy decoding) are used under their official Apache-2.0, MIT, or research release terms.

\paragraph{Grading and model cohorts.}
Predicted answers are normalized (Unicode NFKC normalization, lowercasing, stripping leading answer prefixes, articles, and punctuation, normalizing color synonyms, and converting number words to digits) and matched against the gold answer and accepted aliases by deterministic rules.
Unresolved free-form replies are adjudicated by a frozen text-only Qwen3.8-27B semantic-equivalence judge given the question, gold answer, accepted aliases, and model reply (without image access) to return \emph{equivalent}, \emph{not\_equivalent}, or \emph{unsure}; empty or unparsable lines count as wrong and indeterminate judge outputs ($<0.5\%$) are excluded.
Behavioral analyses on DCI cover all $22$ open-weight VLMs ($1$B--$35$B) in \cref{tab:supp_ability} under a shared fixed question order per scene, while attention readouts, causal interventions, and cross-benchmark masking cover all $14$ models (marked with $\dagger$) supporting custom 4D attention masks and attention-weight readouts.

\begin{table*}[tp]
\centering
\caption{\textbf{Single-question accuracy (\%) by DCI ability category across all 22 open-weight VLMs} ($100$ DCI scenes, $1{,}597$ questions; ordered by overall single-question accuracy as in main Figure~\ref{fig:ledger}a; $\dagger$ marks the $14$ models supporting custom 4D attention masks; \emph{Global} [$199$] denotes the semantic ability category, whereas the $195$ whole-image questions in main Section~3.1 are items without a localized target box).}
\label{tab:supp_ability}
\footnotesize
\renewcommand{\arraystretch}{0.90}
\setlength{\tabcolsep}{0pt}
\begin{tabular*}{\textwidth}{@{\extracolsep{\fill}}rlrrrrrrrrr@{}}
\toprule
Rank & Model & Spatial ($365$) & Color ($304$) & Global ($199$) & Object ($198$) & Count ($194$) & Ordinal ($167$) & Attr.\ ($101$) & Text ($69$) & All ($1{,}597$) \\
\midrule
1 & \raisebox{-0.22ex}{\includegraphics[height=1.65ex]{fig/logos/mistral.pdf}}\hspace{3.2pt}Ministral-3-3B$^{\dagger}$ & 39.7 & 35.2 & 69.1 & 25.8 & 34.0 & 26.9 & 57.9 & 18.8 & 38.9 \\
2 & \raisebox{-0.22ex}{\includegraphics[height=1.65ex]{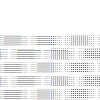}}\hspace{3.2pt}MiniCPM-V-4.6 & 53.0 & 52.6 & 56.3 & 34.4 & 17.5 & 32.9 & 66.0 & 17.4 & 43.9 \\
3 & \raisebox{-0.22ex}{\includegraphics[height=1.65ex]{fig/logos/gemma.pdf}}\hspace{3.2pt}Gemma-4-E2B$^{\dagger}$ & 46.0 & 41.4 & 77.4 & 37.6 & 27.3 & 34.7 & 60.6 & 55.1 & 45.9 \\
4 & \raisebox{-0.22ex}{\includegraphics[height=1.65ex]{fig/logos/mistral.pdf}}\hspace{3.2pt}Ministral-3-8B$^{\dagger}$ & 46.8 & 45.4 & 75.6 & 36.6 & 40.7 & 27.5 & 62.1 & 21.7 & 45.9 \\
5 & \raisebox{-0.22ex}{\includegraphics[height=1.65ex]{fig/logos/internvl.pdf}}\hspace{3.2pt}InternVL3.5-30B-A3B & 56.3 & 65.8 & 72.1 & 48.0 & 29.9 & 43.1 & 61.0 & 24.6 & 53.4 \\
6 & \raisebox{-0.22ex}{\includegraphics[height=1.65ex]{fig/logos/gemma.pdf}}\hspace{3.2pt}Gemma-4-E4B$^{\dagger}$ & 48.9 & 60.9 & 82.3 & 45.6 & 43.8 & 32.9 & 71.3 & 65.2 & 54.8 \\
7 & \raisebox{-0.22ex}{\includegraphics[height=1.65ex]{fig/logos/qwen.pdf}}\hspace{3.2pt}Qwen3-VL-2B$^{\dagger}$ & 52.6 & 66.4 & 82.3 & 60.2 & 45.4 & 37.3 & 78.0 & 63.8 & 59.5 \\
8 & \raisebox{-0.22ex}{\includegraphics[height=1.65ex]{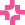}}\hspace{3.2pt}Molmo2-O-7B & 61.8 & 75.3 & 67.8 & 65.7 & 37.1 & 38.3 & 82.2 & 31.9 & 60.2 \\
9 & \raisebox{-0.22ex}{\includegraphics[height=1.65ex]{fig/logos/internvl.pdf}}\hspace{3.2pt}InternVL3-8B$^{\dagger}$ & 51.0 & 73.7 & 88.9 & 66.3 & 45.4 & 43.7 & 83.0 & 50.7 & 62.5 \\
10 & \raisebox{-0.22ex}{\includegraphics[height=1.65ex]{fig/logos/qwen.pdf}}\hspace{3.2pt}Qwen2.5-VL-7B$^{\dagger}$ & 55.3 & 77.3 & 84.9 & 73.1 & 43.3 & 39.5 & 81.0 & 81.2 & 65.0 \\
11 & \raisebox{-0.22ex}{\includegraphics[height=1.65ex]{fig/logos/qwen.pdf}}\hspace{3.2pt}Qwen3.5-4B & 59.7 & 77.3 & 91.4 & 63.2 & 49.5 & 44.9 & 80.0 & 50.0 & 65.5 \\
12 & \raisebox{-0.22ex}{\includegraphics[height=1.65ex]{fig/logos/glm.pdf}}\hspace{3.2pt}GLM-4.6V-Flash$^{\dagger}$ & 60.2 & 77.3 & 88.4 & 73.1 & 46.4 & 45.5 & 85.1 & 70.6 & 67.4 \\
13 & \raisebox{-0.22ex}{\includegraphics[height=1.65ex]{fig/logos/gemma.pdf}}\hspace{3.2pt}Gemma-4-12B$^{\dagger}$ & 61.6 & 77.6 & 93.0 & 65.3 & 57.7 & 49.4 & 89.1 & 52.2 & 68.6 \\
14 & \raisebox{-0.22ex}{\includegraphics[height=1.65ex]{fig/logos/qwen.pdf}}\hspace{3.2pt}Qwen3-VL-4B$^{\dagger}$ & 61.9 & 81.3 & 86.9 & 75.9 & 54.6 & 50.3 & 85.1 & 71.0 & 70.2 \\
15 & \raisebox{-0.22ex}{\includegraphics[height=1.65ex]{fig/logos/qwen.pdf}}\hspace{3.2pt}Qwen3.5-9B & 65.8 & 82.2 & 89.4 & 73.3 & 57.7 & 52.7 & 83.2 & 59.4 & 71.3 \\
16 & \raisebox{-0.22ex}{\includegraphics[height=1.65ex]{fig/logos/qwen.pdf}}\hspace{3.2pt}Qwen3-VL-30B-A3B$^{\dagger}$ & 68.2 & 82.9 & 86.4 & 77.0 & 54.6 & 52.7 & 86.1 & 78.3 & 72.6 \\
17 & \raisebox{-0.22ex}{\includegraphics[height=1.65ex]{fig/logos/qwen.pdf}}\hspace{3.2pt}Qwen3-VL-8B$^{\dagger}$ & 63.0 & 82.2 & 89.9 & 75.1 & 59.3 & 56.9 & 90.1 & 82.6 & 73.0 \\
18 & \raisebox{-0.22ex}{\includegraphics[height=1.65ex]{fig/logos/qwen.pdf}}\hspace{3.2pt}Qwen3.6-35B-A3B & 72.9 & 85.2 & 88.4 & 76.0 & 62.9 & 54.5 & 90.1 & 53.6 & 74.7 \\
19 & \raisebox{-0.22ex}{\includegraphics[height=1.65ex]{fig/logos/qwen.pdf}}\hspace{3.2pt}Qwen3.6-27B & 73.7 & 83.2 & 91.9 & 80.4 & 61.3 & 56.9 & 88.1 & 48.5 & 75.2 \\
20 & \raisebox{-0.22ex}{\includegraphics[height=1.65ex]{fig/logos/qwen.pdf}}\hspace{3.2pt}Qwen3.5-27B & 71.2 & 83.6 & 92.0 & 78.2 & 66.5 & 56.3 & 89.1 & 56.5 & 75.4 \\
21 & \raisebox{-0.22ex}{\includegraphics[height=1.65ex]{fig/logos/qwen.pdf}}\hspace{3.2pt}Qwen3-VL-32B$^{\dagger}$ & 66.0 & 89.1 & 93.4 & 73.6 & 67.5 & 61.4 & 91.1 & 75.4 & 76.5 \\
22 & \raisebox{-0.22ex}{\includegraphics[height=1.65ex]{fig/logos/gemma.pdf}}\hspace{3.2pt}Gemma-4-31B$^{\dagger}$ & 72.3 & 81.3 & 92.0 & 78.2 & 65.5 & 61.7 & 90.1 & 82.6 & 76.8 \\
\bottomrule
\end{tabular*}
\end{table*}

\begin{table*}[tp]
\centering
\caption{\textbf{Per-model DCI co-asking results and Positions~2--16 preceding-context repair and damage rates across 22 open-weight VLMs} ($100$ DCI scenes, percentages; \emph{Rank} $1$--$22$ matches \cref{tab:supp_ability} and main Figure~\ref{fig:ledger}a). Left: single-question accuracy (\emph{Single-Q}) and paired one-pass co-asking repair rate, damage rate, and net accuracy shift (\emph{Net $\Delta$}) across Positions~1--16 under the standard prompt. Middle and right: conditional repair and damage rates across Positions~2--16 comparing same-image questions (\emph{Same}), synthetic text-only questions (\emph{Text-only}), and unseen other-image questions (\emph{Other}) under the preceding-context control prompt, which appends the \emph{N/A} clause to all four arms including its matched single-question baseline (\cref{sec:supp_repair}).}
\label{tab:supp_22models_dci}
\footnotesize
\renewcommand{\arraystretch}{0.90}
\setlength{\tabcolsep}{0pt}
\begin{tabular*}{\textwidth}{@{\extracolsep{\fill}}rlrrrr@{\hspace{10pt}}rrr@{\hspace{10pt}}rrr@{}}
\toprule
 & & \multicolumn{4}{c}{Standard prompt (Pos.\ 1--16)} & \multicolumn{3}{c}{Control Repair (\%)} & \multicolumn{3}{c}{Control Damage (\%)} \\
\cmidrule(lr){3-6} \cmidrule(lr){7-9} \cmidrule(l){10-12}
Rank & Model & Single-Q & Repair & Damage & Net $\Delta$ & Same & Text-only & Other & Same & Text-only & Other \\
\midrule
1 & \raisebox{-0.22ex}{\includegraphics[height=1.65ex]{fig/logos/mistral.pdf}}\hspace{3.2pt}Ministral-3-3B$^{\dagger}$ & 38.9 & 15.4 & 23.8 & $+0.2$ & 13.5 & 8.7 & 5.7 & 25.9 & 29.1 & 56.2 \\
2 & \raisebox{-0.22ex}{\includegraphics[height=1.65ex]{fig/logos/minicpm.pdf}}\hspace{3.2pt}MiniCPM-V-4.6 & 43.9 & 24.8 & 30.8 & $+0.4$ & 22.0 & 20.6 & 23.2 & 30.0 & 40.6 & 34.6 \\
3 & \raisebox{-0.22ex}{\includegraphics[height=1.65ex]{fig/logos/gemma.pdf}}\hspace{3.2pt}Gemma-4-E2B$^{\dagger}$ & 45.9 & 17.0 & 18.1 & $+0.8$ & 21.4 & 17.1 & 21.6 & 16.5 & 14.3 & 18.5 \\
4 & \raisebox{-0.22ex}{\includegraphics[height=1.65ex]{fig/logos/mistral.pdf}}\hspace{3.2pt}Ministral-3-8B$^{\dagger}$ & 45.9 & 16.7 & 22.3 & $-1.3$ & 11.9 & 8.0 & 6.5 & 31.7 & 44.3 & 42.1 \\
5 & \raisebox{-0.22ex}{\includegraphics[height=1.65ex]{fig/logos/internvl.pdf}}\hspace{3.2pt}InternVL3.5-30B-A3B & 53.4 & 29.8 & 14.3 & $+6.3$ & 35.7 & 24.7 & 17.5 & 14.1 & 22.4 & 49.8 \\
6 & \raisebox{-0.22ex}{\includegraphics[height=1.65ex]{fig/logos/gemma.pdf}}\hspace{3.2pt}Gemma-4-E4B$^{\dagger}$ & 54.8 & 16.4 & 16.5 & $-1.6$ & 17.8 & 12.0 & 14.2 & 17.5 & 15.3 & 16.8 \\
7 & \raisebox{-0.22ex}{\includegraphics[height=1.65ex]{fig/logos/qwen.pdf}}\hspace{3.2pt}Qwen3-VL-2B$^{\dagger}$ & 59.5 & 19.1 & 21.6 & $-5.2$ & 23.9 & 20.4 & 11.7 & 22.5 & 25.2 & 57.1 \\
8 & \raisebox{-0.22ex}{\includegraphics[height=1.65ex]{fig/logos/molmo.pdf}}\hspace{3.2pt}Molmo2-O-7B & 60.2 & 24.4 & 22.5 & $-3.8$ & 35.8 & 34.0 & 32.3 & 25.0 & 22.1 & 32.5 \\
9 & \raisebox{-0.22ex}{\includegraphics[height=1.65ex]{fig/logos/internvl.pdf}}\hspace{3.2pt}InternVL3-8B$^{\dagger}$ & 62.5 & 16.8 & 29.2 & $-12.0$ & 16.1 & 12.4 & 13.4 & 31.7 & 23.9 & 38.0 \\
10 & \raisebox{-0.22ex}{\includegraphics[height=1.65ex]{fig/logos/qwen.pdf}}\hspace{3.2pt}Qwen2.5-VL-7B$^{\dagger}$ & 65.0 & 16.3 & 17.5 & $-5.6$ & 19.8 & 13.3 & 8.7 & 15.9 & 19.6 & 44.0 \\
11 & \raisebox{-0.22ex}{\includegraphics[height=1.65ex]{fig/logos/qwen.pdf}}\hspace{3.2pt}Qwen3.5-4B & 65.5 & 19.2 & 14.8 & $-3.1$ & 21.7 & 16.4 & 7.3 & 16.7 & 15.8 & 62.1 \\
12 & \raisebox{-0.22ex}{\includegraphics[height=1.65ex]{fig/logos/glm.pdf}}\hspace{3.2pt}GLM-4.6V-Flash$^{\dagger}$ & 67.4 & 16.6 & 10.9 & $-1.9$ & 17.2 & 11.7 & 9.3 & 11.8 & 12.6 & 25.0 \\
13 & \raisebox{-0.22ex}{\includegraphics[height=1.65ex]{fig/logos/gemma.pdf}}\hspace{3.2pt}Gemma-4-12B$^{\dagger}$ & 68.6 & 16.0 & 8.5 & $-0.8$ & 20.6 & 12.8 & 7.8 & 7.3 & 6.3 & 27.4 \\
14 & \raisebox{-0.22ex}{\includegraphics[height=1.65ex]{fig/logos/qwen.pdf}}\hspace{3.2pt}Qwen3-VL-4B$^{\dagger}$ & 70.2 & 19.2 & 12.7 & $-3.2$ & 20.8 & 12.3 & 14.2 & 12.7 & 15.7 & 22.2 \\
15 & \raisebox{-0.22ex}{\includegraphics[height=1.65ex]{fig/logos/qwen.pdf}}\hspace{3.2pt}Qwen3.5-9B & 71.3 & 21.6 & 11.6 & $-2.1$ & 22.0 & 14.0 & 11.6 & 10.3 & 11.5 & 17.2 \\
16 & \raisebox{-0.22ex}{\includegraphics[height=1.65ex]{fig/logos/qwen.pdf}}\hspace{3.2pt}Qwen3-VL-30B-A3B$^{\dagger}$ & 72.6 & 22.0 & 10.6 & $-1.7$ & 30.0 & 15.9 & 15.2 & 9.2 & 11.7 & 26.3 \\
17 & \raisebox{-0.22ex}{\includegraphics[height=1.65ex]{fig/logos/qwen.pdf}}\hspace{3.2pt}Qwen3-VL-8B$^{\dagger}$ & 73.0 & 18.6 & 13.0 & $-4.5$ & 19.8 & 12.7 & 12.9 & 12.7 & 11.0 & 18.5 \\
18 & \raisebox{-0.22ex}{\includegraphics[height=1.65ex]{fig/logos/qwen.pdf}}\hspace{3.2pt}Qwen3.6-35B-A3B & 74.7 & 23.1 & 10.3 & $-1.8$ & 30.3 & 19.2 & 14.1 & 9.9 & 11.6 & 25.6 \\
19 & \raisebox{-0.22ex}{\includegraphics[height=1.65ex]{fig/logos/qwen.pdf}}\hspace{3.2pt}Qwen3.6-27B & 75.2 & 24.6 & 7.6 & $+0.4$ & 25.0 & 15.3 & 10.9 & 7.0 & 7.0 & 14.3 \\
20 & \raisebox{-0.22ex}{\includegraphics[height=1.65ex]{fig/logos/qwen.pdf}}\hspace{3.2pt}Qwen3.5-27B & 75.4 & 25.3 & 5.5 & $+2.1$ & 25.1 & 12.0 & 12.0 & 5.3 & 6.7 & 15.4 \\
21 & \raisebox{-0.22ex}{\includegraphics[height=1.65ex]{fig/logos/qwen.pdf}}\hspace{3.2pt}Qwen3-VL-32B$^{\dagger}$ & 76.5 & 19.3 & 6.6 & $-0.6$ & 27.1 & 14.9 & 12.8 & 6.2 & 5.5 & 8.2 \\
22 & \raisebox{-0.22ex}{\includegraphics[height=1.65ex]{fig/logos/gemma.pdf}}\hspace{3.2pt}Gemma-4-31B$^{\dagger}$ & 76.8 & 20.3 & 4.7 & $+1.1$ & 21.5 & 11.0 & 11.0 & 5.6 & 6.5 & 11.7 \\
\midrule
--- & Pooled ($22$ models) & 63.6 & 19.9 & 14.1 & $-1.7$ & 22.3 & 15.7 & 13.7 & 14.4 & 15.6 & 28.3 \\
\bottomrule
\end{tabular*}
\end{table*}

\section{Question-Level Analysis and Target Attention}
\label{sec:supp_ledger}\label{sec:supp_legacy}\label{sec:supp_attention}

\begin{figure*}[!t]
\centering
\includegraphics[width=\textwidth]{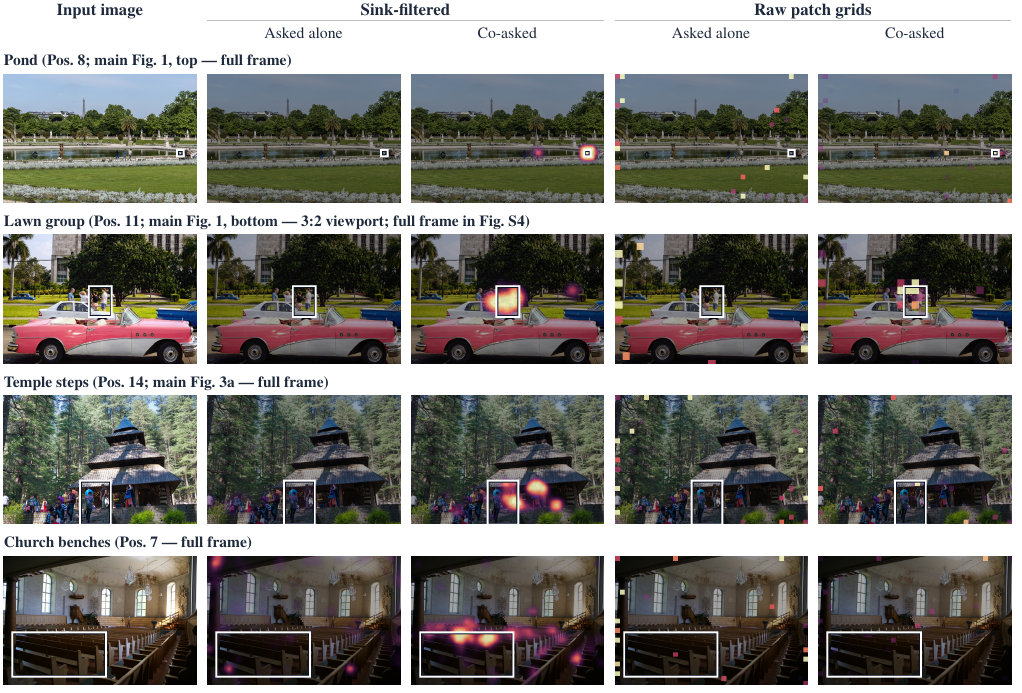}
\caption{\textbf{Sink-filtered and raw ($\sigma=0$) patch-grid controls on Gemma-4-12B.} Symmetric boundary-sink-filtered maps ($\sigma=0.82$ display smoothing, shared per-row 99.5th-percentile ceiling) alongside unsmoothed raw patch grids across four repaired questions. \emph{Pond} (main Figure~\ref{fig:teaser}, top), \emph{Temple steps} (main Figure~\ref{fig:attn_heatmap}a), and \emph{Church benches} show full uncropped $3{:}2$ frames; \emph{Lawn group} (main Figure~\ref{fig:teaser}, bottom) shows its $3{:}2$ viewport, with the full uncropped $2{:}3$ portrait frame in \cref{fig:supp_lawn_fullfield}.}
\label{fig:supp_attn_unsink}
\end{figure*}

\begin{figure}[!t]
\centering
\includegraphics[width=\columnwidth]{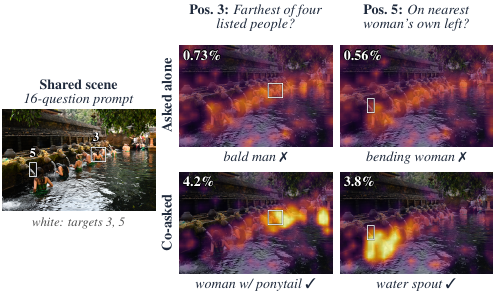}
\caption{\textbf{Additional shared-scene multi-question example on Gemma-4-12B} (Fountain scene, layers $17$, $23$, $29$). Position~3 asks which of four listed people is farthest from the camera, and Position~5 asks which listed object or person is on the nearest woman's own left; raw full-image target-box shares rise from $0.73\%$ to $4.2\%$ and from $0.56\%$ to $3.8\%$.}
\label{fig:supp_fountain}
\end{figure}

\begin{figure}[!t]
\centering
\includegraphics[width=\columnwidth]{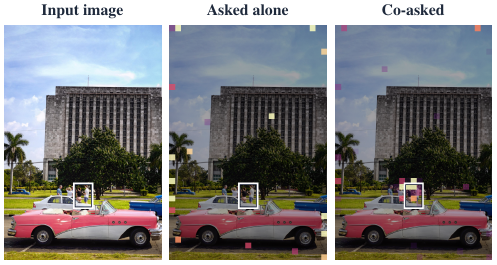}
\caption{\textbf{Full-frame portrait control for the lawn scene in main Figure~\ref{fig:teaser} (bottom).} Uncropped $2{:}3$ raw patch grids ($\sigma=0$) retain boundary sink patches while showing target-box concentration under co-asking; all quantitative target shares precede display cropping.}
\label{fig:supp_lawn_fullfield}
\end{figure}

\paragraph{Per-model DCI breakdowns and target-attention protocol.}
\Cref{tab:supp_ability,tab:supp_22models_dci} report each model's single-question accuracy by ability category and paired co-asking repair, damage, and net accuracy shifts across all $22$ VLMs.
Across the $22$ models, single-question errors are repaired at $15.4\%$--$29.8\%$ regardless of model scale, whereas damage falls from $30.8\%$ on weaker models to $4.7\%$ on stronger ones.
For prefill target localization across the $14$ mask-capable models (\cref{tab:supp_u4_14models}), all four attention conditions (\emph{Alone}, \emph{Co-ask}, \emph{Text-only}, and \emph{Blocked}) use the standard prompt template (without the \emph{N/A} clause) and evaluate Positions~2--16 on the common set of questions admitting both a reference target region and an equal-area non-overlapping random box ($N = 1{,}356$ on all full-attention layers and the $1{,}024$-token sliding-attention layers of Gemma-4-12B and Gemma-4-31B, and $N = 1{,}093$ on the $512$-token local sliding-attention layers of Gemma-4-E2B and Gemma-4-E4B where image tokens remain within the $512$-token window).
Adapting~\cite{zhang2025mllms} without generic-description normalization, each layer's target attention ratio at the final token of $Q_k$ divides the mean target-box attention share by the mean random-box share over that common question set and averages equally across layers (\cref{tab:supp_u4_14models}; dividing by \emph{Alone} yields main Table~\ref{tab:target_lead}).
In qualitative heatmaps, symmetric boundary-sink filtering and Gaussian smoothing are applied for visual display only, while all numeric percentages report raw unsmoothed target-box shares over the full uncropped image; \cref{fig:supp_attn_unsink,fig:supp_fountain,fig:supp_lawn_fullfield} provide sink-filtered versus raw ($\sigma=0$) patch grids (including the full uncropped $3{:}2$ frames for main Figure~\ref{fig:teaser} top and main Figure~\ref{fig:attn_heatmap}a), an additional shared-scene multi-question pair, and the full uncropped $2{:}3$ portrait frame for main Figure~\ref{fig:teaser} bottom.

\begin{table}[!t]
\centering
\caption{\textbf{Raw unnormalized target attention ratios across all 14 models on DCI} (Positions~2--16, final token of each question stem under the standard prompt). Entries report the layer-averaged ratio of mean target-box attention to mean same-size non-overlapping random-box attention; dividing by \emph{Alone} yields main Table~\ref{tab:target_lead}.}
\label{tab:supp_u4_14models}\label{tab:supp_u4_13models}
\footnotesize\renewcommand{\arraystretch}{0.92}\setlength{\tabcolsep}{0pt}
\begin{tabular*}{\columnwidth}{@{\extracolsep{\fill}}lrrrr@{}}
\toprule
Model & Alone & Co-ask & Text-only & Blocked \\
\midrule
\raisebox{-0.22ex}{\includegraphics[height=1.65ex]{fig/logos/qwen.pdf}}\hspace{3.2pt}Qwen3-VL-2B & 2.556 & \textbf{2.997} & 2.401 & 2.687 \\
\raisebox{-0.22ex}{\includegraphics[height=1.65ex]{fig/logos/qwen.pdf}}\hspace{3.2pt}Qwen3-VL-4B & 2.475 & \textbf{2.675} & 2.392 & 2.398 \\
\raisebox{-0.22ex}{\includegraphics[height=1.65ex]{fig/logos/qwen.pdf}}\hspace{3.2pt}Qwen3-VL-8B & 2.191 & \textbf{2.351} & 2.114 & 2.138 \\
\raisebox{-0.22ex}{\includegraphics[height=1.65ex]{fig/logos/qwen.pdf}}\hspace{3.2pt}Qwen3-VL-30B-A3B & 4.006 & \textbf{4.468} & 3.638 & 3.855 \\
\raisebox{-0.22ex}{\includegraphics[height=1.65ex]{fig/logos/qwen.pdf}}\hspace{3.2pt}Qwen3-VL-32B & 2.167 & \textbf{2.397} & 2.107 & 2.178 \\
\raisebox{-0.22ex}{\includegraphics[height=1.65ex]{fig/logos/qwen.pdf}}\hspace{3.2pt}Qwen2.5-VL-7B & 2.676 & \textbf{3.302} & 2.841 & 2.921 \\
\raisebox{-0.22ex}{\includegraphics[height=1.65ex]{fig/logos/gemma.pdf}}\hspace{3.2pt}Gemma-4-E2B & 1.754 & \textbf{2.725} & 2.377 & 2.690 \\
\raisebox{-0.22ex}{\includegraphics[height=1.65ex]{fig/logos/gemma.pdf}}\hspace{3.2pt}Gemma-4-E4B & 2.100 & \textbf{3.169} & 2.882 & 3.068 \\
\raisebox{-0.22ex}{\includegraphics[height=1.65ex]{fig/logos/gemma.pdf}}\hspace{3.2pt}Gemma-4-12B & 1.763 & \textbf{2.251} & 2.089 & 2.158 \\
\raisebox{-0.22ex}{\includegraphics[height=1.65ex]{fig/logos/gemma.pdf}}\hspace{3.2pt}Gemma-4-31B & 1.865 & \textbf{2.499} & 2.367 & 2.399 \\
\raisebox{-0.22ex}{\includegraphics[height=1.65ex]{fig/logos/mistral.pdf}}\hspace{3.2pt}Ministral-3-3B & 2.052 & \textbf{2.077} & 1.977 & 1.997 \\
\raisebox{-0.22ex}{\includegraphics[height=1.65ex]{fig/logos/mistral.pdf}}\hspace{3.2pt}Ministral-3-8B & 2.067 & \textbf{2.226} & 2.009 & 2.103 \\
\raisebox{-0.22ex}{\includegraphics[height=1.65ex]{fig/logos/internvl.pdf}}\hspace{3.2pt}InternVL3-8B & 2.697 & \textbf{3.099} & 2.683 & 2.874 \\
\raisebox{-0.22ex}{\includegraphics[height=1.65ex]{fig/logos/glm.pdf}}\hspace{3.2pt}GLM-4.6V-Flash & 2.574 & \textbf{2.971} & 2.545 & 2.752 \\
\bottomrule
\end{tabular*}
\end{table}

\section{Cross-Question Visual Grounding}
\label{sec:supp_repair}

\begin{table}[!t]
\centering
\caption{\textbf{Pooled DCI repair rate (\%) across question positions by preceding context and regional target-box overlap} (shared control prompt across $22$ VLMs; top block covers all $1{,}597$ questions as in main Figure~\ref{fig:ledger}c; bottom block splits the $1{,}402$ regional questions by $\text{IoU}$ with earlier target boxes as in main Figure~\ref{fig:ledger}e,f; regional text-only repair across Positions~2--16 is $15.0\%$).}
\label{tab:supp_nonoverlap}
\footnotesize
\renewcommand{\arraystretch}{0.92}
\setlength{\tabcolsep}{0pt}
\begin{tabular*}{\columnwidth}{@{\extracolsep{\fill}}lrrrrr@{}}
\toprule
 & \multicolumn{5}{c}{Question position} \\
\cmidrule(l){2-6}
Setting / Cohort & 1 & 2--5 & 6--10 & 11--16 & 2--16 \\
\midrule
Same-image ($22$ VLMs)  & 11.1 & 20.3 & 21.8 & 23.9 & 22.3 \\
Text-only ($22$ VLMs)   & 10.0 & 17.3 & 16.1 & 14.5 & 15.7 \\
Other-image ($22$ VLMs) & 8.3 & 15.7 & 13.2 & 13.1 & 13.7 \\
\midrule
22 VLMs ($\text{IoU}<0.1$)    & --- & 18.6 & 21.2 & 22.5 & 20.9 \\
22 VLMs ($\text{IoU}\ge 0.1$) & --- & 25.0 & 20.3 & 24.7 & 23.6 \\
\bottomrule
\end{tabular*}
\end{table}

\paragraph{Preceding-context and regional-overlap controls.}
In main Section~4.1 (\cref{fig:supp_benchmark_examples}b), synthetic text-only questions match DCI stems in character length and numbering using image-free arithmetic, alphabetical sorting, and word-count tasks, while other-image questions sample real DCI questions from a different, unseen scene while viewing the target image.
Because other-image questions query objects absent from the displayed scene, all four arms of this comparison (\emph{Single-Q}, \emph{Same}, \emph{Text-only}, and \emph{Other}) append \emph{``If a question cannot be answered, answer N/A.''} to the prompt header.
As detailed per model in \cref{tab:supp_22models_dci} and across positions in \cref{tab:supp_nonoverlap}, same-image repair climbs monotonically with position even when earlier target boxes do not overlap the current target ($\text{IoU} < 0.1$: $18.6\% \to 22.5\%$), whereas text-only and other-image controls peak at Positions~2--5 and decline thereafter.

\begin{figure*}[tp]
\centering
\includegraphics[width=\textwidth]{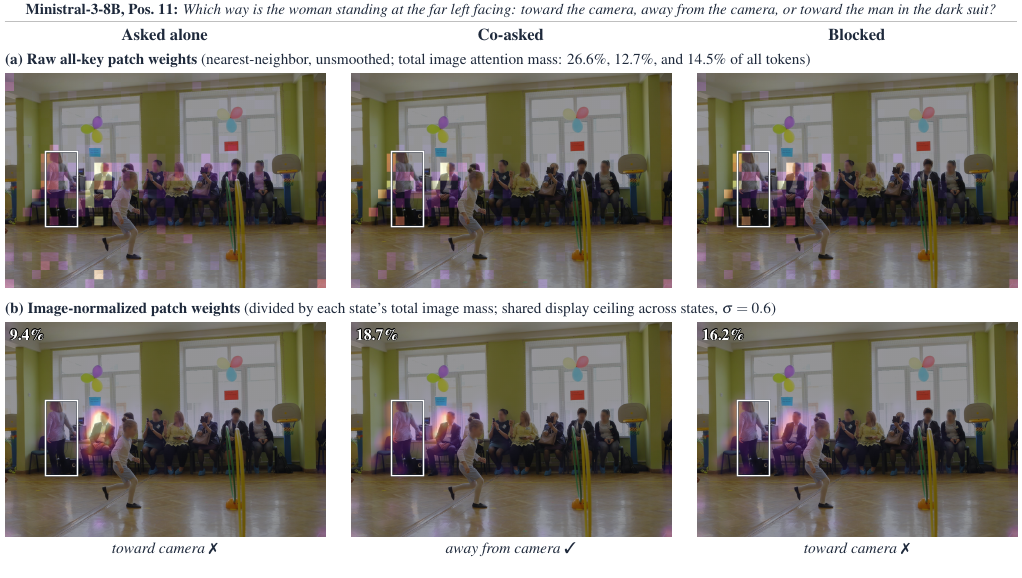}
\caption{\textbf{Raw all-key and image-normalized patch-attention views on a repaired Ministral-3-8B gymnasium question} (Position~11, layers $9$--$18$). (a) Raw patch weights retain the all-key softmax denominator across image and text tokens (whose total image attention mass shifts from $26.6\%$ when asked alone to $12.7\%$ and $14.5\%$ in the $16$-question prompt), displayed with nearest-neighbor patches and no smoothing. (b) Dividing each map by its total image attention mass under a shared display ceiling ($\sigma = 0.6$) isolates spatial redistribution within the image: unsmoothed target-box share rises from $9.4\%$ to $18.7\%$ under co-asking and recedes to $16.2\%$ when $V \to Q_{<k}$ is blocked.}
\label{fig:supp_states_native}\label{fig:supp_native_states}
\end{figure*}

\paragraph{All-layer and windowed blocking of earlier visual reading.}
Under all-layer blocking ($V \nrightarrow Q_{<k}$ in main Section~4.2), Positions~2--16 accuracy drops by $-1.77$ percentage points pooled ($N = 20{,}797$: $-1.06$ at Positions~2--5 [$N = 5{,}546$], $-1.41$ at Positions~6--10 [$N = 6{,}954$], and $-2.56$ at Positions~11--16 [$N = 8{,}297$]), erasing $24.6\%$ of repaired answers pooled ($19.3\%$--$36.9\%$ per model; \cref{tab:supp_window_sweep}).
For the layer-windowed sweep in main Section~4.3 (\cref{tab:supp_window_sweep}), each model's $L$ layers ($0 \dots L-1$) are covered by $12$ overlapping contiguous windows of nominal width $w \approx L/6$ ($13$ windows on $64$-layer Qwen3-VL-32B, with the final window aligned to $L-1$).
At Positions~2--16, we evaluate fixed-seed matched strata of up to $80$ repaired items ($R$: wrong alone, right under our method; $N = 80$ on $12$ models, $77$ on Gemma-4-31B, and $64$ on Qwen3-VL-32B, using all available repaired items) and $80$ control items answered right in both settings ($C$).
Blocking a single early-to-middle reading window removes $13.8\%$--$33.8\%$ of repaired answers while leaving $C$ largely unchanged ($0.0\%$--$6.3\%$) and deep-layer windows in the upper third of the network near zero ($0.0\%$--$5.0\%$).
Each model's subsequent refocusing window on $Q_k$ (\cref{tab:supp_window_sweep}, rightmost column) is the window with the largest drop in $Q_k$'s final-token target attention ratio under all-layer $V \nrightarrow Q_{<k}$ blocking, and \cref{fig:supp_states_native} compares raw all-key and image-normalized patch maps across \emph{Alone}, \emph{Co-asked}, and \emph{Blocked}.

\begin{table*}[tp]
\centering
\caption{\textbf{All-layer and layer-windowed blocking of earlier questions' image attention ($V \nrightarrow Q_{<k}$) and subsequent target refocusing across all 14 models on DCI} (Positions~2--16, inclusive $0$-based layer intervals $a$--$b$). Columns report all-layer repaired-answer loss (\emph{All $R$ lost}, underlying main Section~4.2; pooled $24.6\%$, $368/1{,}494$), total layers $L$ with nominal window width ($w$), per-stratum sample size $N$, peak $V \nrightarrow Q_{<k}$ reading window, percentage of repaired items ($R$) and doubly correct items ($C$) lost in that window, deep-layer window range (upper third of layers) and $R$ losses, and the $Q_k$ refocusing window (with single-layer attention-ratio peak and window-start lag).}
\label{tab:supp_window_sweep}\label{tab:layer_windows}
\footnotesize
\renewcommand{\arraystretch}{0.90}
\setlength{\tabcolsep}{0pt}
\begin{tabular*}{\textwidth}{@{\extracolsep{\fill}}lrcrcrrccc@{}}
\toprule
Model & All $R$ lost (\%) & $L$ ($w$) & $N$ & Peak win. & Win.\ $R$ (\%) & Win.\ $C$ (\%) & Deep layers & Deep $R$ (\%) & Refocus win.\ (peak, lag) \\
\midrule
\raisebox{-0.22ex}{\includegraphics[height=1.65ex]{fig/logos/qwen.pdf}}\hspace{3.2pt}Qwen3-VL-2B          & 21.2 & $28$ ($5$)  & 80 & $6$--$10$             & 16.3 & 6.3 & $17$--$27$ & 0.0--2.5 & $13$--$17$ ($16$, $+7$) \\
\raisebox{-0.22ex}{\includegraphics[height=1.65ex]{fig/logos/qwen.pdf}}\hspace{3.2pt}Qwen3-VL-4B          & 28.0 & $36$ ($6$)  & 80 & $11$--$16$            & 25.0 & 3.8 & $22$--$35$ & 1.3--5.0 & $19$--$24$ ($21$, $+8$) \\
\raisebox{-0.22ex}{\includegraphics[height=1.65ex]{fig/logos/qwen.pdf}}\hspace{3.2pt}Qwen3-VL-8B          & 28.8 & $36$ ($6$)  & 80 & $12$--$17$            & 17.5 & 6.3 & $24$--$35$ & 0.0--1.3 & $18$--$23$ ($21$, $+6$) \\
\raisebox{-0.22ex}{\includegraphics[height=1.65ex]{fig/logos/qwen.pdf}}\hspace{3.2pt}Qwen3-VL-30B-A3B     & 22.7 & $48$ ($8$)  & 80 & $15$--$22$            & 21.3 & 0.0 & $33$--$47$ & 2.5--5.0 & $25$--$32$ ($30$, $+10$) \\
\raisebox{-0.22ex}{\includegraphics[height=1.65ex]{fig/logos/qwen.pdf}}\hspace{3.2pt}Qwen3-VL-32B         & 28.1 & $64$ ($11$) & 64 & $35$--$45$            & 18.8 & 4.7 & $50$--$63$ & 0.0--1.6 & $40$--$50$ ($48$, $+5$) \\
\raisebox{-0.22ex}{\includegraphics[height=1.65ex]{fig/logos/qwen.pdf}}\hspace{3.2pt}Qwen2.5-VL-7B        & 19.3 & $28$ ($5$)  & 80 & $8$--$12$             & 13.8 & 1.3 & $19$--$27$ & 0.0--3.8 & $17$--$21$ ($19$, $+9$) \\
\raisebox{-0.22ex}{\includegraphics[height=1.65ex]{fig/logos/gemma.pdf}}\hspace{3.2pt}Gemma-4-E2B         & 24.8 & $35$ ($6$)  & 80 & $5$--$10$             & 18.8 & 1.3 & $21$--$34$ & 0.0--0.0 & $16$--$21$ ($19$, $+11$) \\
\raisebox{-0.22ex}{\includegraphics[height=1.65ex]{fig/logos/gemma.pdf}}\hspace{3.2pt}Gemma-4-E4B         & 23.1 & $42$ ($7$)  & 80 & $10$--$16$            & 17.5 & 5.0 & $25$--$41$ & 0.0--0.0 & $19$--$25$ ($23$, $+9$) \\
\raisebox{-0.22ex}{\includegraphics[height=1.65ex]{fig/logos/gemma.pdf}}\hspace{3.2pt}Gemma-4-12B         & 36.9 & $48$ ($8$)  & 80 & $16$--$23$            & 23.8 & 1.3 & $28$--$47$ & 1.3--5.0 & $24$--$31$ ($29$, $+8$) \\
\raisebox{-0.22ex}{\includegraphics[height=1.65ex]{fig/logos/gemma.pdf}}\hspace{3.2pt}Gemma-4-31B         & 36.4 & $60$ ($10$) & 77 & $20$--$29$            & 33.8 & 3.9 & $35$--$59$ & 0.0--2.6 & $30$--$39$ ($35$, $+10$) \\
\raisebox{-0.22ex}{\includegraphics[height=1.65ex]{fig/logos/mistral.pdf}}\hspace{3.2pt}Ministral-3-3B    & 19.5 & $26$ ($4$)  & 80 & $10$--$13$            & 13.8 & 5.0 & $16$--$25$ & 0.0--3.8 & $16$--$19$ ($16$, $+6$) \\
\raisebox{-0.22ex}{\includegraphics[height=1.65ex]{fig/logos/mistral.pdf}}\hspace{3.2pt}Ministral-3-8B    & 22.3 & $34$ ($6$)  & 80 & $12$--$17$            & 20.0 & 6.3 & $21$--$33$ & 0.0--3.8 & $15$--$20$ ($17$, $+3$) \\
\raisebox{-0.22ex}{\includegraphics[height=1.65ex]{fig/logos/internvl.pdf}}\hspace{3.2pt}InternVL3-8B     & 22.6 & $28$ ($5$)  & 80 & $13$--$17$            & 18.8 & 1.3 & $19$--$27$ & 5.0--5.0 & $17$--$21$ ($19$, $+4$) \\
\raisebox{-0.22ex}{\includegraphics[height=1.65ex]{fig/logos/glm.pdf}}\hspace{3.2pt}GLM-4.6V-Flash        & 24.4 & $40$ ($7$)  & 80 & $0$--$6$ / $12$--$18$ & 15.0 / 13.8 & 1.3 / 3.8 & $24$--$39$ & 0.0--2.5 & $18$--$24$ ($20$, $+6$) \\
\bottomrule
\end{tabular*}
\end{table*}

\section{Answer-Time Masking and Position Re-Indexing}
\label{sec:supp_method}

In main Section~5.1, paired masked runs (\emph{Co-Asking}, \emph{Ours}, and \emph{Blocked}) decode each question $Q_k$ individually while holding earlier answers $A_{<k}$ fixed to the model's one-pass co-asking output so altered earlier answers cannot cascade across positions.
When decoding Question~$k$ under \emph{Read Together, Answer Apart} (\emph{Ours}) and \emph{Blocked}, the post-prompt segment---comprising the chat-template assistant header, fixed prior answers $A_{<k}$, and current answer tokens $A_k$---masks attention to later question stems ($Q_{j > k}$), and $A_k$ additionally masks attention to earlier question stems ($Q_{j < k}$), keeping shared prefill representations of $V$ and $Q_k$ intact while retaining $A_{<k}$ for output-line numbering.%
\begin{table*}[tp]
\centering
\caption{\textbf{Main-order paired flip counts and three-order DCI robustness across all 14 models.} \emph{Main-order $N$ ($+/-$)} gives the four-setting common valid sample size and answers newly right/wrong under \emph{Ours} vs.\ \emph{Co-Ask} underlying main Table~\ref{tab:main_dci}. Remaining columns pool three random question orders per scene (\emph{Order~1} from the main paper plus two independent random permutations \emph{Orders~2--3}; $N = 66{,}554$ total, $N_{2\text{--}16} = 62{,}385$ for $\Delta\text{Blk}_{2\text{--}16} = \text{Blocked} - \text{Ours}$); \emph{1-Q} reports full-set single-question accuracy as in \cref{tab:supp_ability}, while \emph{Co-Ask} and \emph{Ours} report paired fixed-prefix accuracies on the common valid set.}
\label{tab:supp_dci_14models}\label{tab:supp_seed_gains}\label{tab:supp_knockout_13models}
\footnotesize\renewcommand{\arraystretch}{0.92}\setlength{\tabcolsep}{0pt}
\begin{tabular*}{\textwidth}{@{\extracolsep{\fill}}lrrrrrrrrr@{}}
\toprule
Model & Main-order $N$ ($+/-$) & $N_{\text{3-ord}}$ & 1-Q & Co-Ask & \textbf{Ours} & Pooled $\Delta$ & $+ / -$ & Ord.\ $1 / 2 / 3$ $\Delta$ & $\Delta\text{Blk}_{2\text{--}16}$ \\
\midrule
\raisebox{-0.22ex}{\includegraphics[height=1.65ex]{fig/logos/qwen.pdf}}\hspace{3.2pt}Qwen2.5-VL-7B & 1,593 (125/50) & 4,778 & 65.0 & 60.9 & \textbf{64.2} & $+3.26$ & 319/163 & $+4.71$/$+2.20$/$+2.89$ & $-1.45$ \\
\raisebox{-0.22ex}{\includegraphics[height=1.65ex]{fig/logos/internvl.pdf}}\hspace{3.2pt}InternVL3-8B & 1,588 (118/57) & 4,767 & 62.5 & 50.5 & \textbf{53.7} & $+3.19$ & 347/195 & $+3.84$/$+4.15$/$+1.57$ & $-0.54$ \\
\raisebox{-0.22ex}{\includegraphics[height=1.65ex]{fig/logos/mistral.pdf}}\hspace{3.2pt}Ministral-3-3B & 1,556 (124/68) & 4,677 & 38.9 & 39.3 & \textbf{42.4} & $+3.12$ & 365/219 & $+3.60$/$+1.99$/$+3.77$ & $-0.57$ \\
\raisebox{-0.22ex}{\includegraphics[height=1.65ex]{fig/logos/qwen.pdf}}\hspace{3.2pt}Qwen3-VL-2B & 1,590 (130/66) & 4,768 & 59.5 & 55.2 & \textbf{57.8} & $+2.58$ & 326/203 & $+4.03$/$+1.32$/$+2.39$ & $-3.07$ \\
\raisebox{-0.22ex}{\includegraphics[height=1.65ex]{fig/logos/glm.pdf}}\hspace{3.2pt}GLM-4.6V-Flash & 1,588 (79/53) & 4,769 & 67.4 & 65.1 & \textbf{67.4} & $+2.31$ & 251/141 & $+1.64$/$+3.27$/$+2.01$ & $-0.54$ \\
\raisebox{-0.22ex}{\includegraphics[height=1.65ex]{fig/logos/qwen.pdf}}\hspace{3.2pt}Qwen3-VL-30B-A3B & 1,594 (75/45) & 4,779 & 72.6 & 71.4 & \textbf{73.3} & $+1.93$ & 239/147 & $+1.88$/$+1.76$/$+2.14$ & $-2.14$ \\
\raisebox{-0.22ex}{\includegraphics[height=1.65ex]{fig/logos/qwen.pdf}}\hspace{3.2pt}Qwen3-VL-8B & 1,592 (77/47) & 4,774 & 73.0 & 68.9 & \textbf{70.5} & $+1.63$ & 208/130 & $+1.88$/$+2.07$/$+0.94$ & $-2.79$ \\
\raisebox{-0.22ex}{\includegraphics[height=1.65ex]{fig/logos/mistral.pdf}}\hspace{3.2pt}Ministral-3-8B & 1,567 (99/63) & 4,701 & 45.9 & 45.5 & \textbf{46.9} & $+1.36$ & 303/239 & $+2.30$/$+0.76$/$+1.02$ & $-1.93$ \\
\raisebox{-0.22ex}{\includegraphics[height=1.65ex]{fig/logos/qwen.pdf}}\hspace{3.2pt}Qwen3-VL-4B & 1,588 (77/57) & 4,758 & 70.2 & 67.3 & \textbf{68.4} & $+1.16$ & 224/169 & $+1.26$/$+1.51$/$+0.69$ & $-2.87$ \\
\raisebox{-0.22ex}{\includegraphics[height=1.65ex]{fig/logos/qwen.pdf}}\hspace{3.2pt}Qwen3-VL-32B & 1,582 (44/35) & 4,754 & 76.5 & 76.1 & \textbf{77.0} & $+0.86$ & 147/106 & $+0.57$/$+0.44$/$+1.58$ & $-3.57$ \\
\raisebox{-0.22ex}{\includegraphics[height=1.65ex]{fig/logos/gemma.pdf}}\hspace{3.2pt}Gemma-4-12B & 1,593 (47/24) & 4,780 & 68.6 & 68.4 & \textbf{68.9} & $+0.46$ & 127/105 & $+1.44$/$-0.38$/$+0.31$ & $-0.51$ \\
\raisebox{-0.22ex}{\includegraphics[height=1.65ex]{fig/logos/gemma.pdf}}\hspace{3.2pt}Gemma-4-E4B & 1,581 (73/79) & 4,742 & 54.8 & 53.1 & \textbf{53.6} & $+0.42$ & 236/216 & $-0.38$/$+0.82$/$+0.82$ & $-0.61$ \\
\raisebox{-0.22ex}{\includegraphics[height=1.65ex]{fig/logos/gemma.pdf}}\hspace{3.2pt}Gemma-4-31B & 1,593 (21/20) & 4,778 & 76.8 & \textbf{78.8} & 78.7 & $-0.19$ & 67/76 & $+0.06$/$-0.13$/$-0.50$ & $-1.92$ \\
\raisebox{-0.22ex}{\includegraphics[height=1.65ex]{fig/logos/gemma.pdf}}\hspace{3.2pt}Gemma-4-E2B & 1,576 (93/119) & 4,729 & 45.9 & \textbf{46.7} & 45.1 & $-1.65$ & 285/363 & $-1.65$/$-2.22$/$-1.08$ & $-0.27$ \\
\midrule
Pooled ($14$ models) & 22,181 (1,182/783) & 66,554 & 62.8 & 60.6 & \textbf{62.0} & $+1.46$ & 3,444/2,472 & $+1.80$/$+1.28$/$+1.30$ & $-1.63$ \\
\bottomrule\end{tabular*}
\end{table*}%
\ Because masking $Q_{j > k}$ removes the intervening stems between $Q_k$ and the assistant turn, rotary position re-indexing (\emph{Ours} and \emph{Blocked}, whereas \emph{Co-Asking} retains native linear positions) assigns the post-prompt segment consecutive rotary position IDs immediately following $Q_k$'s trailing line-break slot---accounting across all six model series for whether the tokenizer merges the trailing newline into the final stem token (\eg, ``?\textbackslash n'' on Qwen2.5-VL, Qwen3-VL, Ministral-3, InternVL3, and GLM-4.6V-Flash) or emits a standalone newline token (Gemma-4), and shifting all three $(t,h,w)$ text coordinates by the same scalar offset under 3D M-RoPE (Qwen2.5-VL, Qwen3-VL, and GLM-4.6V-Flash).

\section{Question-Order Robustness and Candidate Margins}
\label{sec:supp_three_orders}

\paragraph{Question-order robustness.}
To verify that neither cross-question visual help nor answer-time stem isolation depends on a single permutation, \cref{tab:supp_dci_14models} evaluates all $14$ models across three random question orders per DCI scene (the main-paper order \emph{Order~1} plus two additional independent permutations \emph{Orders~2--3}, totaling $N = 66{,}554$ paired observations: $4{,}169$ at Position~1 and $62{,}385$ at Positions~2--16).
Across all three orders, isolating answer decoding nearly halves Position-1 damaged answers ($339 \to 174$) while preserving Positions~2--16 repaired answers ($4{,}307 \to 4{,}411$) and cutting Positions~2--16 damaged answers ($5{,}590 \to 4{,}789$).
\emph{Read Together, Answer Apart} improves pooled accuracy in every individual order ($+1.80$, $+1.28$, and $+1.30$ percentage points; $+1.46$ pooled across $12$ of $14$ models), while blocking $V \to Q_{<k}$ across all layers lowers Positions~2--16 accuracy in all $14$ models ($-1.63$ points pooled, deepening from $-0.69$ at Positions~2--5 to $-1.56$ at Positions~6--10 and $-2.31$ at Positions~11--16).

\paragraph{Closed-candidate log-likelihood margins.}
On the $1{,}046$ closed-candidate DCI questions ($414$ multiple-choice, $376$ color-answer, and $256$ integer-answer items across ability categories), we score each candidate answer without free-form generation by summing token log-probabilities immediately after the numbered answer prefix across all $14$ models (\cref{tab:supp_cand_margin}).
Mean ground-truth minus strongest-distractor log-likelihood margin increases under \emph{Ours} relative to co-asking on $13$ of $14$ models ($+0.07$ to $+0.86$) and drops under \emph{Blocked} relative to \emph{Ours} on $13$ of $14$ models ($-0.02$ to $-1.42$), with Gemma-4-E2B as the sole exception in both comparisons.

\begin{table}[!tb]
\centering
\caption{\textbf{Closed-candidate log-likelihood margins across all 14 models on DCI} ($N = 1{,}046$ closed-candidate questions: ground-truth candidate log-likelihood minus strongest-distractor log-likelihood; $\Delta_{\text{Ours}} = \text{Ours} - \text{Co-ask}$ and $\Delta_{\text{Blk}} = \text{Blocked} - \text{Ours}$ are computed prior to two-decimal rounding).}
\label{tab:supp_cand_margin}
\footnotesize\renewcommand{\arraystretch}{0.90}\setlength{\tabcolsep}{0pt}
\begin{tabular*}{\columnwidth}{@{\extracolsep{\fill}}lrrrrrr@{}}
\toprule
Model & 1-Q & Co-ask & \textbf{Ours} & Blk & $\Delta_{\text{Ours}}$ & $\Delta_{\text{Blk}}$ \\
\midrule
\raisebox{-0.22ex}{\includegraphics[height=1.65ex]{fig/logos/qwen.pdf}}\hspace{3.2pt}Qwen3-VL-2B & 1.28 & 0.47 & \textbf{0.91} & 0.42 & $+0.44$ & $-0.49$ \\
\raisebox{-0.22ex}{\includegraphics[height=1.65ex]{fig/logos/qwen.pdf}}\hspace{3.2pt}Qwen3-VL-4B & 3.98 & 3.14 & \textbf{4.01} & 2.66 & $+0.86$ & $-1.34$ \\
\raisebox{-0.22ex}{\includegraphics[height=1.65ex]{fig/logos/qwen.pdf}}\hspace{3.2pt}Qwen3-VL-8B & 3.41 & 3.45 & \textbf{4.09} & 2.67 & $+0.64$ & $-1.42$ \\
\raisebox{-0.22ex}{\includegraphics[height=1.65ex]{fig/logos/qwen.pdf}}\hspace{3.2pt}Qwen3-VL-30B-A3B & 2.77 & 2.56 & \textbf{3.00} & 2.26 & $+0.45$ & $-0.74$ \\
\raisebox{-0.22ex}{\includegraphics[height=1.65ex]{fig/logos/qwen.pdf}}\hspace{3.2pt}Qwen3-VL-32B & 6.22 & 5.45 & \textbf{6.14} & 4.82 & $+0.68$ & $-1.32$ \\
\raisebox{-0.22ex}{\includegraphics[height=1.65ex]{fig/logos/qwen.pdf}}\hspace{3.2pt}Qwen2.5-VL-7B & 0.97 & 0.56 & \textbf{0.75} & 0.56 & $+0.20$ & $-0.20$ \\
\raisebox{-0.22ex}{\includegraphics[height=1.65ex]{fig/logos/gemma.pdf}}\hspace{3.2pt}Gemma-4-E2B & $-0.45$ & $\mathbf{-0.02}$ & $-0.44$ & $-0.42$ & $-0.42$ & $+0.02$ \\
\raisebox{-0.22ex}{\includegraphics[height=1.65ex]{fig/logos/gemma.pdf}}\hspace{3.2pt}Gemma-4-E4B & 0.65 & 0.42 & \textbf{0.51} & 0.33 & $+0.10$ & $-0.19$ \\
\raisebox{-0.22ex}{\includegraphics[height=1.65ex]{fig/logos/gemma.pdf}}\hspace{3.2pt}Gemma-4-12B & 3.45 & 3.81 & \textbf{3.91} & 3.40 & $+0.10$ & $-0.52$ \\
\raisebox{-0.22ex}{\includegraphics[height=1.65ex]{fig/logos/gemma.pdf}}\hspace{3.2pt}Gemma-4-31B & 8.04 & 7.35 & \textbf{7.68} & 6.41 & $+0.32$ & $-1.27$ \\
\raisebox{-0.22ex}{\includegraphics[height=1.65ex]{fig/logos/mistral.pdf}}\hspace{3.2pt}Ministral-3-3B & $-0.58$ & $-0.48$ & $\mathbf{-0.36}$ & $-0.38$ & $+0.12$ & $-0.02$ \\
\raisebox{-0.22ex}{\includegraphics[height=1.65ex]{fig/logos/mistral.pdf}}\hspace{3.2pt}Ministral-3-8B & $-0.24$ & $-0.39$ & $\mathbf{-0.32}$ & $-0.38$ & $+0.07$ & $-0.06$ \\
\raisebox{-0.22ex}{\includegraphics[height=1.65ex]{fig/logos/internvl.pdf}}\hspace{3.2pt}InternVL3-8B & 1.75 & 0.54 & \textbf{0.97} & 0.73 & $+0.43$ & $-0.24$ \\
\raisebox{-0.22ex}{\includegraphics[height=1.65ex]{fig/logos/glm.pdf}}\hspace{3.2pt}GLM-4.6V-Flash & 2.63 & 2.16 & \textbf{2.60} & 2.22 & $+0.44$ & $-0.38$ \\
\bottomrule
\end{tabular*}
\end{table}

\section{Additional Benchmarks}
\label{sec:supp_additional}\label{sec:supp_benchmarks}

\begin{figure*}[!t]
\centering
\includegraphics[width=0.95\textwidth]{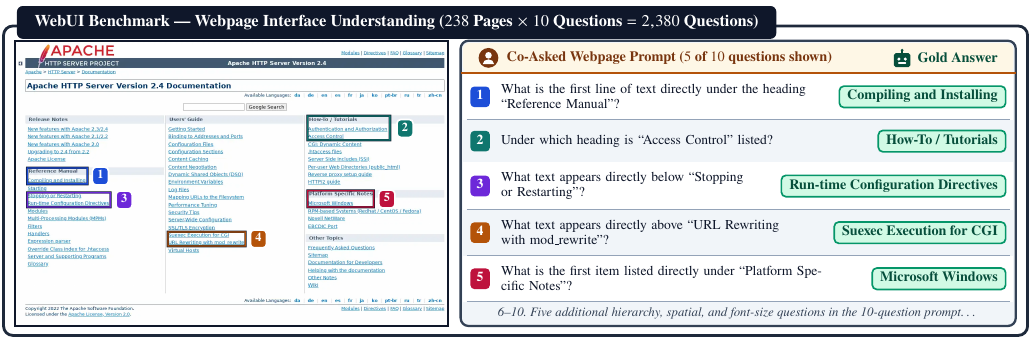}
\vspace{-3pt}
\caption{\textbf{Representative WebUI benchmark example} ($238$ webpage screenshots, $10$ co-asked questions per page). Color-coded region markers highlight the DOM elements queried by Questions~1--5 alongside verbatim question stems and gold answers.}
\label{fig:supp_webui_example}
\vspace{-4pt}
\end{figure*}

\begin{figure*}[!t]
\centering
\includegraphics[width=0.95\textwidth]{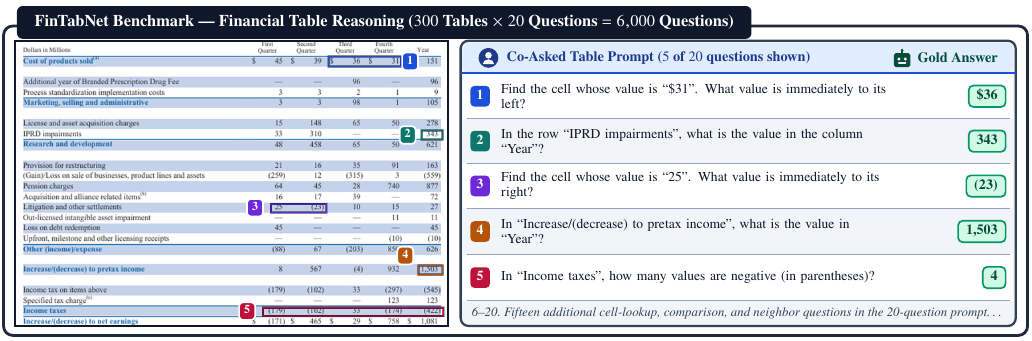}
\vspace{-3pt}
\caption{\textbf{Representative FinTabNet benchmark example} ($300$ financial tables, $20$ co-asked questions per table). Color-coded boxes mark the cells and rows queried by Questions~1--5 alongside question stems and gold answers.}
\label{fig:supp_fintab_example}
\vspace{-4pt}
\end{figure*}

\begin{figure*}[!t]
\centering
\includegraphics[width=0.95\textwidth]{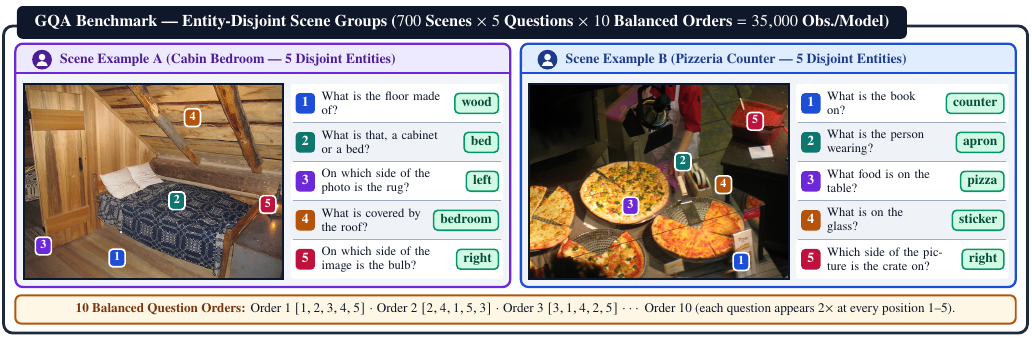}
\vspace{-3pt}
\caption{\textbf{Representative GQA benchmark examples} ($700$ scenes, $5$ entity-disjoint questions per scene rotated across $10$ balanced question orders). Two scenes show the five disjoint queried entities, verbatim stems, gold answers, and balanced position rotation.}
\label{fig:supp_gqa_example}
\vspace{-2pt}
\end{figure*}

\Cref{fig:supp_webui_example,fig:supp_fintab_example,fig:supp_gqa_example} show scene and prompt examples for WebUI, FinTabNet, and GQA, and \cref{tab:supp_cross_native_repair,tab:supp_cross_dataset} report per-model co-asking repair rates and four-setting accuracies on WebUI ($N = 33{,}283$ paired observations), FinTabNet ($N = 83{,}855$), and GQA ($N = 484{,}798$ across $10$ balanced orders), plus Positions~2--10 earlier-image blocking ($\Delta_{\text{Blk}} = \text{Blocked} - \text{Ours}$, $N = 29{,}952$) on WebUI.
Across all three benchmarks, co-asking repair rises substantially after Position~1 (\cref{tab:supp_cross_native_repair}), and blocking $V \to Q_{<k}$ on WebUI lowers Positions~2--10 accuracy in all $14$ models ($-2.9$ points pooled).
Isolating answer decoding yields the largest gains where co-asking incurs substantial cross-stem damage---improving $10$ of $14$ models on WebUI ($+2.3$ points pooled, up to $+11.4$) and $8$ of $14$ models on FinTabNet (up to $+5.3$; the two smallest Gemma-4 models account for the $-0.3$ pooled shift, whereas the other $12$ models gain $+0.9$ pooled)---while remaining neutral ($0.0$ pooled) on $5$-question entity-disjoint GQA prompts where co-asking incurs minimal net damage.

\vfill\break
\noindent\begin{minipage}{\columnwidth}
\centering
\captionof{table}{\textbf{One-pass co-asking repair rate (\%) across WebUI, FinTabNet, and GQA} (per-model and position-pooled across all $14$ models; \emph{Early / Middle / Late} denote Positions~2--4 / 5--8 / 9--10 on WebUI, 2--8 / 9--12 / 13--20 on FinTabNet, and 2 / 3 / 4--5 on GQA).}
\label{tab:supp_cross_native_repair}\label{tab:supp_cross_position}
\vspace{2pt}
\footnotesize\renewcommand{\arraystretch}{0.83}\setlength{\tabcolsep}{0pt}
\begin{tabular*}{\columnwidth}{@{\extracolsep{\fill}}lrrr@{}}
\toprule
Model / Position bin & WebUI & FinTabNet & GQA \\
\midrule
\raisebox{-0.22ex}{\includegraphics[height=1.65ex]{fig/logos/qwen.pdf}}\hspace{3.2pt}Qwen3-VL-2B & 12.9 & 14.7 & 16.6 \\
\raisebox{-0.22ex}{\includegraphics[height=1.65ex]{fig/logos/qwen.pdf}}\hspace{3.2pt}Qwen3-VL-4B & 19.9 & 17.0 & 15.6 \\
\raisebox{-0.22ex}{\includegraphics[height=1.65ex]{fig/logos/qwen.pdf}}\hspace{3.2pt}Qwen3-VL-8B & 19.1 & 22.6 & 11.8 \\
\raisebox{-0.22ex}{\includegraphics[height=1.65ex]{fig/logos/qwen.pdf}}\hspace{3.2pt}Qwen3-VL-30B-A3B & 23.8 & 18.7 & 16.4 \\
\raisebox{-0.22ex}{\includegraphics[height=1.65ex]{fig/logos/qwen.pdf}}\hspace{3.2pt}Qwen3-VL-32B & 20.4 & 23.8 & 12.0 \\
\raisebox{-0.22ex}{\includegraphics[height=1.65ex]{fig/logos/qwen.pdf}}\hspace{3.2pt}Qwen2.5-VL-7B & 21.6 & 21.7 & 12.6 \\
\raisebox{-0.22ex}{\includegraphics[height=1.65ex]{fig/logos/gemma.pdf}}\hspace{3.2pt}Gemma-4-E2B & 16.4 & 20.1 & 18.0 \\
\raisebox{-0.22ex}{\includegraphics[height=1.65ex]{fig/logos/gemma.pdf}}\hspace{3.2pt}Gemma-4-E4B & 15.9 & 17.4 & 11.5 \\
\raisebox{-0.22ex}{\includegraphics[height=1.65ex]{fig/logos/gemma.pdf}}\hspace{3.2pt}Gemma-4-12B & 16.8 & 35.8 & 9.8 \\
\raisebox{-0.22ex}{\includegraphics[height=1.65ex]{fig/logos/gemma.pdf}}\hspace{3.2pt}Gemma-4-31B & 16.4 & 46.8 & 7.8 \\
\raisebox{-0.22ex}{\includegraphics[height=1.65ex]{fig/logos/mistral.pdf}}\hspace{3.2pt}Ministral-3-3B & 14.6 & 19.2 & 18.1 \\
\raisebox{-0.22ex}{\includegraphics[height=1.65ex]{fig/logos/mistral.pdf}}\hspace{3.2pt}Ministral-3-8B & 16.0 & 22.6 & 14.5 \\
\raisebox{-0.22ex}{\includegraphics[height=1.65ex]{fig/logos/internvl.pdf}}\hspace{3.2pt}InternVL3-8B & 14.7 & 14.8 & 19.0 \\
\raisebox{-0.22ex}{\includegraphics[height=1.65ex]{fig/logos/glm.pdf}}\hspace{3.2pt}GLM-4.6V-Flash & 25.9 & 21.7 & 12.5 \\
\midrule
Pooled ($14$ models, all pos.) & 17.3 & 20.2 & 14.2 \\
\quad Position 1 & 12.3 & 12.4 & 9.7 \\
\quad Early positions & 16.3 & 19.5 & 14.6 \\
\quad Middle positions & 18.8 & 20.8 & 15.4 \\
\quad Late positions & 18.3 & 21.4 & 15.7 \\
\bottomrule
\end{tabular*}
\end{minipage}

\begin{table*}[!t]
\centering
\caption{\textbf{Cross-dataset accuracy (\%) across all 14 models on WebUI, FinTabNet, and GQA.} Columns report single-question querying (1-Q), fixed-prefix per-question co-asking (Co), \emph{Read Together, Answer Apart} (Ours), and $\Delta_{\text{Ours}} = \text{Ours} - \text{Co}$ on each benchmark's common valid set, plus Positions~2--10 earlier-image blocking ($\Delta_{\text{Blk}} = \text{Blocked} - \text{Ours}$) on WebUI (boldface marks the unrounded higher of \emph{Co} and \emph{Ours} per model, and both on the GQA pooled tie; model-level improvement counts in main Table~\ref{tab:cross_bench_summary} likewise compare unrounded accuracies).}
\label{tab:supp_cross_dataset}
\footnotesize
\renewcommand{\arraystretch}{0.90}
\setlength{\tabcolsep}{0pt}
\begin{tabular*}{\textwidth}{@{\extracolsep{\fill}}lrrrrr@{\hspace{12pt}}rrrr@{\hspace{12pt}}rrrr@{}}
\toprule
 & \multicolumn{5}{c}{WebUI ($10$ questions / image)} & \multicolumn{4}{c}{FinTabNet ($20$ questions / table)} & \multicolumn{4}{c}{GQA ($5$ questions / image)} \\
\cmidrule(lr){2-6} \cmidrule(lr){7-10} \cmidrule(l){11-14}
Model & 1-Q & Co & \textbf{Ours} & $\Delta_{\text{Ours}}$ & $\Delta_{\text{Blk}}$ & 1-Q & Co & \textbf{Ours} & $\Delta_{\text{Ours}}$ & 1-Q & Co & \textbf{Ours} & $\Delta_{\text{Ours}}$ \\
\midrule
\raisebox{-0.22ex}{\includegraphics[height=1.65ex]{fig/logos/qwen.pdf}}\hspace{3.2pt}Qwen3-VL-2B & 64.9 & 46.9 & \textbf{57.7} & $+10.8$ & $-5.0$ & 16.2 & 18.5 & \textbf{18.8} & $+0.3$ & 68.4 & \textbf{67.2} & 67.1 & $-0.2$ \\
\raisebox{-0.22ex}{\includegraphics[height=1.65ex]{fig/logos/qwen.pdf}}\hspace{3.2pt}Qwen3-VL-4B & 77.6 & 72.9 & \textbf{74.2} & $+1.2$ & $-4.2$ & 36.3 & 35.1 & \textbf{35.2} & $+0.1$ & 69.0 & \textbf{69.7} & 69.3 & $-0.4$ \\
\raisebox{-0.22ex}{\includegraphics[height=1.65ex]{fig/logos/qwen.pdf}}\hspace{3.2pt}Qwen3-VL-8B & 79.5 & 77.4 & \textbf{78.7} & $+1.3$ & $-3.9$ & 37.8 & \textbf{40.0} & 39.8 & $-0.2$ & 70.9 & 70.9 & \textbf{70.9} & $0.0$ \\
\raisebox{-0.22ex}{\includegraphics[height=1.65ex]{fig/logos/qwen.pdf}}\hspace{3.2pt}Qwen3-VL-30B-A3B & 80.6 & 78.1 & \textbf{79.6} & $+1.6$ & $-1.0$ & 47.2 & 42.5 & \textbf{45.5} & $+2.9$ & 72.9 & \textbf{74.2} & 74.2 & $0.0$ \\
\raisebox{-0.22ex}{\includegraphics[height=1.65ex]{fig/logos/qwen.pdf}}\hspace{3.2pt}Qwen3-VL-32B & 84.8 & \textbf{86.0} & 85.6 & $-0.4$ & $-0.7$ & 54.4 & 55.0 & \textbf{55.9} & $+0.9$ & 67.2 & \textbf{68.4} & 67.8 & $-0.6$ \\
\raisebox{-0.22ex}{\includegraphics[height=1.65ex]{fig/logos/qwen.pdf}}\hspace{3.2pt}Qwen2.5-VL-7B & 78.1 & 73.1 & \textbf{75.4} & $+2.3$ & $-2.9$ & 40.6 & 41.7 & \textbf{44.6} & $+3.0$ & 67.3 & 65.8 & \textbf{66.8} & $+1.0$ \\
\raisebox{-0.22ex}{\includegraphics[height=1.65ex]{fig/logos/gemma.pdf}}\hspace{3.2pt}Gemma-4-E2B & 61.8 & 53.1 & \textbf{55.4} & $+2.2$ & $-2.3$ & 52.4 & \textbf{45.3} & 32.8 & $-12.5$ & 53.2 & 54.4 & \textbf{54.8} & $+0.4$ \\
\raisebox{-0.22ex}{\includegraphics[height=1.65ex]{fig/logos/gemma.pdf}}\hspace{3.2pt}Gemma-4-E4B & 72.7 & 67.0 & \textbf{68.4} & $+1.4$ & $-4.1$ & 65.3 & \textbf{56.5} & 54.0 & $-2.5$ & 59.5 & \textbf{59.1} & 59.1 & $0.0$ \\
\raisebox{-0.22ex}{\includegraphics[height=1.65ex]{fig/logos/gemma.pdf}}\hspace{3.2pt}Gemma-4-12B & 83.4 & \textbf{84.5} & 84.1 & $-0.5$ & $-0.9$ & 81.6 & \textbf{83.4} & 82.6 & $-0.8$ & 65.7 & \textbf{65.8} & 65.4 & $-0.5$ \\
\raisebox{-0.22ex}{\includegraphics[height=1.65ex]{fig/logos/gemma.pdf}}\hspace{3.2pt}Gemma-4-31B & 87.4 & \textbf{88.4} & 88.2 & $-0.2$ & $-1.3$ & 89.2 & \textbf{91.5} & 90.6 & $-0.9$ & 67.4 & \textbf{67.4} & 67.3 & $-0.1$ \\
\raisebox{-0.22ex}{\includegraphics[height=1.65ex]{fig/logos/mistral.pdf}}\hspace{3.2pt}Ministral-3-3B & 46.3 & 44.4 & \textbf{44.9} & $+0.5$ & $-4.7$ & 31.0 & 34.2 & \textbf{34.4} & $+0.2$ & 48.2 & \textbf{51.2} & 50.3 & $-0.9$ \\
\raisebox{-0.22ex}{\includegraphics[height=1.65ex]{fig/logos/mistral.pdf}}\hspace{3.2pt}Ministral-3-8B & 51.0 & \textbf{50.1} & 50.0 & $0.0$ & $-5.1$ & 33.4 & 39.1 & \textbf{39.6} & $+0.5$ & 49.4 & \textbf{51.3} & 51.0 & $-0.2$ \\
\raisebox{-0.22ex}{\includegraphics[height=1.65ex]{fig/logos/internvl.pdf}}\hspace{3.2pt}InternVL3-8B & 70.4 & 44.1 & \textbf{55.4} & $+11.4$ & $-1.4$ & 37.2 & 29.4 & \textbf{34.7} & $+5.3$ & 69.8 & 66.2 & \textbf{67.3} & $+1.1$ \\
\raisebox{-0.22ex}{\includegraphics[height=1.65ex]{fig/logos/glm.pdf}}\hspace{3.2pt}GLM-4.6V-Flash & 78.4 & 76.0 & \textbf{77.1} & $+1.1$ & $-2.9$ & 34.6 & \textbf{40.1} & 40.0 & $-0.1$ & 70.6 & 70.4 & \textbf{70.7} & $+0.3$ \\
\midrule
Pooled ($14$ models) & 72.6 & 67.3 & \textbf{69.6} & $+2.3$ & $-2.9$ & 46.9 & \textbf{46.6} & 46.3 & $-0.3$ & 64.3 & \textbf{64.4} & \textbf{64.4} & $0.0$ \\
\bottomrule
\end{tabular*}
\end{table*}

\end{document}